\RequirePackage{fix-cm}
\documentclass[final,smallextended]{svjour3}

\smartqed

\usepackage{graphicx}

 \usepackage{subfig}
 \usepackage{epsfig}
 \usepackage[utf8]{inputenc}
 \usepackage{amsmath,amssymb,amsfonts,nomencl, mathtools} 
 \usepackage{booktabs}
 \usepackage{algorithmicx}
 \usepackage{amsmath}

 \usepackage{amstext,amsmath,amssymb}
 \usepackage{algorithm}
 \usepackage{hyperref}
 \usepackage{algpseudocode}
 \usepackage{xcolor}
 \usepackage{natbib}
 \usepackage{multirow}
 \usepackage{color,xcolor}
 \newcommand{\tabincell}[2]{\begin{tabular}{@{}#1@{}}#2\end{tabular}}
 
 \usepackage{tikz}
\usepackage{pgfplots}

\usepackage{appendix}

\graphicspath{{figs/}}

\usepackage[markup=none,commentmarkup=footnote]{changes}
\usepackage{color}
\definechangesauthor[color=red]{YS}

\definecolor{marine}{RGB}{0,32,96}
\definecolor{navy}{RGB}{0,0,128}
\definecolor{maroon}{RGB}{128,0,0}
\definecolor{olivegreen}{RGB}{85,107,47}
\definecolor{gray}{RGB}{102,102,102}
\definecolor{green}{RGB}{131,198,210}
\definecolor{blue}{rgb}{0, 0.4470, 0.7410}
\definecolor{skyblue}{rgb}{0.3010, 0.7450, 0.9330}
\definechangesauthor[color=maroon]{AE}
\definechangesauthor[color=blue]{XL}

\begin{document}


\title{Vehicle Re-identification Based on Dual Distance Center Loss}

\author{Zhijun Hu \and Yong Xu \and Jie Wen  \and Lilei Sun \and RAJA S P
}

\institute{Zhijun Hu (Corresponding Author) \and Lilei Sun  \at
         College of Computer Science and Technology, Guizhou University, Guiyang, 550025, China \\
        \email{huzhijun@mailbox.gxnu.edu.cn; sunlileisun@163.com}           
        \and
        Yong Xu \and Jie Wen \at
        Shenzhen Key Laboratory of Visual Object Detection and Recognition, Shenzhen, China \\
        \email{laterfall@hit.edu.cn; jiewen\_pr@126.com} 
        \and
        RAJA S P \at
        Associate Professor, Department of Computer Science and Engineering, 
        Vel Tech Rangarajan Dr. Sagunthala R\&D, Chennai, Tamilnadu, India \\
        \email{drspraja@veltech.edu.in}
}

\date{Received: date / Accepted: date}



\maketitle

\section*{Abstract}
	Recently, deep learning has been widely used in the field of vehicle re-identification. When training a deep model, softmax loss is usually used as a supervision tool. However, the softmax loss performs well for closed-set tasks, but not very well for open-set tasks. In this paper, we sum up five shortcomings of center loss and solved all of them by proposing a dual distance center loss (DDCL). Especially we solve the shortcoming that center loss must combine with the softmax loss to supervise training the model, which provides us with a new perspective to examine the center loss. In addition, we verify the inconsistency between the proposed DDCL and softmax loss in the feature space, which makes the center loss no longer be limited by the softmax loss in the feature space after removing the softmax loss. To be specifically, we add the Pearson distance on the basis of the Euclidean distance to the same center, which makes all features of the same class be confined to the intersection of a hypersphere and a hypercube in the feature space. The proposed Pearson distance strengthens the intra-class compactness of the center loss and enhances the generalization ability of center loss. Moreover, by designing a Euclidean distance threshold between all center pairs, which not only strengthens the inter-class separability of center loss, but also makes the center loss (or DDCL) works well without the combination of softmax loss. We apply DDCL in the field of vehicle re-identification named VeRi-776 dataset and VehicleID dataset. And in order to verify its good generalization ability, we also verify it in two datasets commonly used in the field of person re-identification named MSMT17 dataset and Market1501 dataset. The experimental results of the proposed DDCL exceed that of the softmax loss in all the four datasets we used. In the two datasets with larger number of training IDs, the experimental results of the DDCL exceed that of the combination of the softmax loss and the original center loss , indicating that DDCL performs better on large datasets, which verifies the superiority of DDCL.

\section{Introduction}
\label{intro}
\ \ \ \ \ \ Due to the good structure \citep{szegedy2015going,huang2017densely,he2016deep} and the excellent learning method \citep{kingma2014adam,ruder2016overview}, the convolutional neural network (CNN) is widely used in traffic speed prediction and autonomous driving and other urban traffic tasks. It is also well done in vehicle re-identification task \cite{zhuge2020attribute}. The purpose of vehicle re-identification is to find the same vehicle from non-overlapping camera scenes \cite{meng2020parsing}.

\begin{figure}
	\includegraphics[width=1\textwidth]{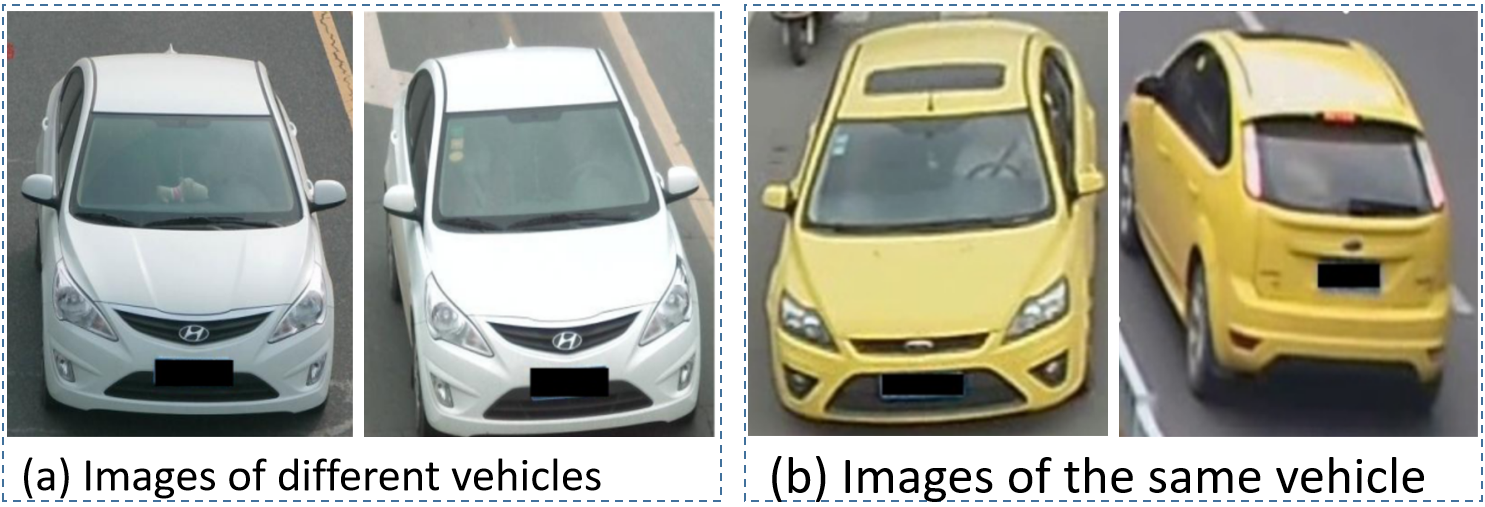}
	\caption{Vehicle images of the same vehicle look different, while images of different vehicles may look similar.}
	\label{vehicle}       
\end{figure}

The two challenges of vehicle re-identification are shown in Fig. \ref{vehicle}. The first challenge is referred as the huge intra-class differences which mean that under different conditions of occlusion, light, perspective, distance, etc., images of the same vehicle appear to be very different. The second challenge is referred as the small inter-class differences, that is, different vehicles with the same color and model produced by the same manufacturer may look very similar \citep{meng2020parsing,bai2018group,kuma2019vehicle,zhou2018aware}. Therefore, reducing the intra-class differences and increasing the inter-class differences are the most important points for vehicle re-identification. In \citep{liu2016deep,chu2019vehicle,lou2019embedding,tang2017multi,zhang2017improving}, scholars narrowed the distance of the same vehicle images through metric learning, and  increased the distance between different vehicle images at the same time. In \citep{liu2016large,khorramshahi2019dual,he2019part,liu2018ram,wang2017orientation} scholars segmented vehicle parts in different ways which is used to make full use of the local features in the special part of vehicles to decrease the distance of images of the same vehicle.

For tasks such as action recognition and image classification, in which the datasets used are closed-sets (the testing set and the training set have the same labels) , the labels control the performance in the testing process \citep{wen2016discriminative,liu2017sphereface}, softmax loss can do these classification tasks well. But for recognition tasks such as face recognition, person re-identification and vehicle re-identification, in which the datasets used are open-sets (the labels of the testing set are not included in the training set),  the features extracted by the model control the performance in the testing process, but these features are only intermediate products of the model, the softmax loss only focuses on improving the classification performance, does not focus on the feature separability and softmax loss will not perform well in the open-set tasks. The loss function plays a vital role in the training process and different loss functions will lead to different features.

Center loss \citep{wen2016discriminative} is widely used in open-set tasks. In this paper, we list a series of shortcomings about center loss \cite{wen2016discriminative}, and try our best to improve them, and then apply the improved center loss to vehicle re-identification task. Center loss \citep{wen2016discriminative} hopes to reduce the intra-class differences by narrowing the Euclidean distance between features and their centers. However, due to the insufficient participation of center parameters in the training process, center loss is not well done in reducing intra-class differences. Meanwhile, center loss does not pay any attention to increase the inter-class differences.

We believe that the center loss \cite{wen2016discriminative} has the following five shortcomings. First, the generalization ability is poor, because it is very sensitive to different parameter initialization methods. Second, it must be used with the combination of softmax loss, because once it is used  alone, the model will make all features and centers nearly to a same point during training process. Third, it only takes the consideration of the intra-class compactness. It doesn't consider the inter-class separability since it only designs the supervision mechanism by which features approach to their centers, without the supervision mechanism of pushing away the distance between features and centers of other classes. But triplet loss \citep{schroff2015facenet,hoffer2015deep} and contrastive loss \citep{chopra2005learning,hadsell2006dimensionality} design the two mechanisms simultaneously. Fourth, it must add additional center parameters, which seriously occupies system memory if the training set has a large number of training IDs, so that the program is unable to run. Fifth, during the training process, due to insufficient participation of the center parameters, it is difficult to obtain the optimal centers. In the training of intra-class compactness , the center parameters are only related to the vehicle features of its own class, and have no relationship with that of other classes, the number of optimization times is very limited, in a training epoch, it is no bigger than the maximum of the total number of vehicle images of this class. While in the training of inter-class separability aspect, the participation of each center parameter is 0, because the center loss \cite{wen2016discriminative} does not design inter-class separability mechanism. Compared with triplet loss and contrastive loss, in which all the parameters are optimized to narrow the distance between all positive pairs input and increase the distance between all negative pairs input simultaneously.

In order to improve the first shortcoming above mentioned, considering the Euclidean distance and the Pearson distance are two completely different distances in the feature space, they are highly complementary, center loss \cite{wen2016discriminative} only takes the consideration of the Euclidean distance, so we add the Pearson distance to the same center. By combining the two distances together, features of the same class are confined to the intersection of a hypersphere and a hypercone in the feature space, which strengthens the intra-class compactness and enhances the generalization ability of the center loss. Here we need to point out that since the definition of Pearson distance is related to correlation coefficient, and correlation coefficient is the result of cosine distance between two vectors after de averaging. The Pearson distance in this paper can be replaced by cosine distance. 

In order to improve the last four shortcomings, we design a center isolation loss ($\mathcal{L}_{CI}$) which is used to increase the distance between all center pairs, we call the loss function combined by Euclidean distance center loss, Pearson center loss and center isolation loss as dual distance center loss (DDCL). Because the softmax loss and the proposed DDCL are inconsistent in the feature space, we removed the softmax loss. Through the proposed $\mathcal{L}_{CI}$, we improve the following four shortcomings simultaneously. First, the $\mathcal{L}_{CI}$ increases the distance of all center pairs by setting a threshold, so that all centers will not be trained to the same point, which makes the softmax loss can be removed. Second, through this $\mathcal{L}_{CI}$, it is obvious that we achieve the purpose of increasing the inter-class separability, because we let the distances between all center pairs greater than a threshold. Third, although the center loss adds additional new center parameters, the number of center parameters is the same as that of the weight parameters of the FC layer (see Fig. \ref{network}), and further more, the FC layer has bias parameters, after removing the softmax loss, the FC layer is removed too. So we do not increase the parameters of the whole network, on the contrary, we reduced the network parameters. Fourth, by using our design of $\mathcal{L}_{CI}$, we calculate the distances between all class centers, we maximize the participation of center parameters. So $\mathcal{L}_{CI}$ improves the above four problems. In addition, in order to optimize the center parameters more smoothly, we design a small sample inhibition factor for $\mathcal{L}_{CI}$, so that when the number of center pairs whose distance are less than the given threshold is too small, the learning speed of the parameters of these center pairs is reduced.

In summary, the main contributions of this paper are as follows:

1) Based on the Euclidean distance center loss in \cite{wen2016discriminative}, we add the Pearson distance loss to the same centers to enhance the intra-class compactness. And we verify that the Euclidean distance center loss in \cite{wen2016discriminative} does not have the generalization ability. Pearson distance is added that can enhance the generalization ability of the center loss.

2) We propose a dual distance center loss (DDCL) by designing a center isolation loss ($\mathcal{L}_{CI}$) between all center pairs, which can make DDCL work well without the softmax loss. We verify that the softmax loss and center loss are inconsistent in the feature space, the proposed DDCL gets rid of the constraint of the softmax loss in the feature space, which allows us to examine the center loss  from a new perspective.

3) Extensive experiments are done on two widely used datasets in the field of vehicle re-identification VeRi-776 datasets and vehicleID datasets  to verify the feasibility of DDCL. In addition, we also verify our method in two widely used datasets on person re-identification task named MSMT17 dataset and Market1501 dataset. All these experiments have verified the generalization ability of DDCL .

The rest of this paper is organized as follows. In section \ref{sec:2}, we summarize some research works related to our proposed method. In section \ref{sec:3}, we introduce the proposed methods. In section \ref{sec:4}, we use a lots of experiments to verify the superiority of our method. In section \ref{sec:5}, we summarize the full papers.

\section{Related work}
\label{sec:2}
\subsection{Vehicle re-identification}
\label{sec:2}
\ \ \ \ Since \cite{liu2016large} pointed out the importance of vehicle re-identification in public transport, vehicle re-identification was widely concerned, and scholars began to improve vehicle re-identification methods from different perspectives. In the beginning, some scholars extracted hand features for vehicle re-identification. \cite{zapletal2016vehicle} used color histogram, histograms of oriented gradients and method of linear regression to solve the re-identification problem. \cite{tang2017multi} proposed a multi-modal metric learning structure, which integrated the deep feature, LBP feature map and a Bag-of-Word-based Color Name (BoW-CN) feature and other hand features into an end-to-end network. However, due to the shallow structure of alexnet network, the re-identification accuracy was low. The hand feature extraction was relatively difficult, and the re-identification accuracy was not high. Later, researchers turned to the deep learning method. \cite{liu2016adeep} used the spatio-temporal information contained in the datasets and proposed to treat vehicle re-identification task as two specific progressive search processes: coarse-to-fine search in feature space and near-to-distant search in real-world surveillance environment. This model was relatively complex and was not an end-to-end implement. Moreover, not all datasets provide spatio-temporal information. Then, scholars began to find some efficient networks, which greatly improved the performance of vehicle re-identification. Based on the VGG\_CNN\_M\_1024 model proposed by \cite{chatfield2014return}, \cite{liu2016deep} designed a two branches network, which could directly measure the relative distance of two vehicle images. \cite{shen2017learning} and \cite{zhou2018vehicle,zhou2018vehicle2} used the LSTM network to conduct a long term vehicle information memory for vehicle re-identification task. \\cite{zhou2017cross,zhou2018vehicle2} and \cite{wu2018joint} used the Generative Adversarial Networks (GANs) to generate vehicle images of different views to supplement the training set.  In \cite{hu2017deep,zhang2017improving,cui2017vehicle,xu2018framework,zhou2018vehicle}, scholars began to use deeper and wider networks such as VGG and ResNet for vehicle re-identification, greatly improved the experimental accuracy. 

Recently, some scholars extracted local features for vehicle re-identification.  \cite{liu2016large} used multimodal features, including visual features, license plate, camera and other information for vehicle re-identification. \cite{liu2018ram} and \cite{he2019part} segmented vehicle part region to extract local features. \cite{khorramshahi2019dual} and \cite{wang2017orientation} defined 20 key points and 8 orientations on the vehicle to extract local features. The limitation of local features is that with the change of perspective, local features do not exist stably. Compared with the instability of local features, metric learning is a stable and effective method. The goal of metric learning is to either shorten the distance between images of the same vehicle or increase the distance between images of different vehicle, or both. \cite{guo2019two} proposed a two-level attention network based on Multi-grain Ranking loss (TAMR) to learn an effective feature embedding method for vehicle re-identification. In \cite{bai2018group,kuma2019vehicle,zhang2017improving}, the triplet loss was applied to vehicle re-identification, and the triplet loss function was improved to shorten the distance between positive pairs and push away the distance between negative pairs. \cite{liu2016deep} designed a deep relative distance learning (DRDL) module, in which a coupled clusters loss and a mixed difference network structure were introduced. Inspired by the way humans recognize objects, \cite{chu2019vehicle} proposed a metric learning method based on viewpoint perception. The designed loss function makes the feature distance of the same vehicle in different views smaller than that of different vehicles in the same view point, which reduces the intra-class compactness and increases the inter-class separability.

\subsection{Improved loss function}
\label{sec:22}
\ \ \ \ Loss function plays a role of supervising and restraining parameters during training process. At present, there are many improved loss functions  are available including clustering loss, contrastive loss, margin loss, improved softmax loss, triplet loss and center loss, etc. The coupled clusters loss designed by \cite{liu2016deep} combined the clustering method and triplet loss to form a new loss.  \cite{zhou2018vehicle2} designed a kind of clustering loss for person re re-identification. The training batch size is 256, including randomly selected 16 classes and  16 randomly selected images for each class. The contrastive loss was proposed by \cite{chopra2005learning,hadsell2006dimensionality}, the purpose was to use the Siamese network to compute the distance between two feature pairs, and the distance  can be directly used to determine whether a pair of face images belongs to the same person. \cite{cheng2019modified} pointed out that contrastive loss had two shortcomings, and established a model which combined the modified contrast loss and a Bayesian model to improve the two shortcomings. The purpose of margin loss was to increase the gap between different classes by setting a threshold, which increases the distance between classes and reduces the distance within classes. The marginal loss proposed by \cite{deng2017marginal} combined softmax loss to obtain more discriminative deep features by focusing on marginal samples while minimizing intra-class variance and maximizing inter-class distance. \cite{li2020boosting} proposed an adaptive margin principle, the samples in the feature embedding space were separated from the similar classes, and an additional marginal loss related to tasks was developed to better distinguish different types of samples. \cite{wu2020rotation} redefined the quantization error in angular space and decomposed it into class error and individual error. On this basis, a rotation consistency margin loss is proposed to minimize individual error. Softmax loss was very effective for learning the label mapping, but it did not take consideration of the separability of features. \cite{liu2016large2} improved the softmax loss by designed a multiplicative angular margin to suppress the model parameters, and perfected it in SphereFace \cite{liu2017sphereface}, by normalizing each column of the weight parameter of the linear classification layer, and setting the bias to 0. CosFace \cite{wang2018cosface} designed an additive cosine margin based on SphereFace \cite{liu2017sphereface}, which greatly improved the feature distinction, because addition was more controllable than multiplication. On the basis of CosFace \cite{wang2018cosface}, ArcFace \cite{deng2019arcface} changed the additive cosine margin into the additive angular margin, which again improved the feature distinction. According to the idea of curriculum learning, CurricularFace \cite{huang2020curricularface} proposed to train simple samples in the early stage, and to train hard samples in the later stage. The triplet loss proposed by \cite{schroff2015facenet} input an anchor sample, a positive sample and a negative sample into the network simultaneously. The network shortened the distance between anchor and positive sample, and increased the distance between anchor sample and negative sample. Scholars have proposed many improved versions of triplet loss \cite{hoffer2015deep,cheng2016person,taha2019defense,chen2017beyond}. The center loss was proposed by \cite{wen2016discriminative}. By randomly initializing the center parameters, and minimizing the distance between features and their centers, the center parameters were optimized. The \cite{wen2016discriminative} improved the center loss in \cite{wen2019comprehensive}, by sharing the parameters of the classification layer and the center parameters, which reduced the intra-class distance. \cite{cai2018island} proposed a method which increases the  cosine value of the angle between center pairs to increase the inter-class separability of center loss.

\section{Proposed methods}
\label{sec:3}
\ \ \ \ The center loss \cite{wen2016discriminative} only considered the Euclidean distance center. Based on Euclidean distance center, we add Pearson distance to the same center , which increases the intra-class compactness of features. Then, we restrain the distances between all center pairs bigger than a predetermined threshold, which increases the inter-class separability of features. The model used in this paper is shown in Figure 2.

\begin{figure}
	\includegraphics[width=1\textwidth]{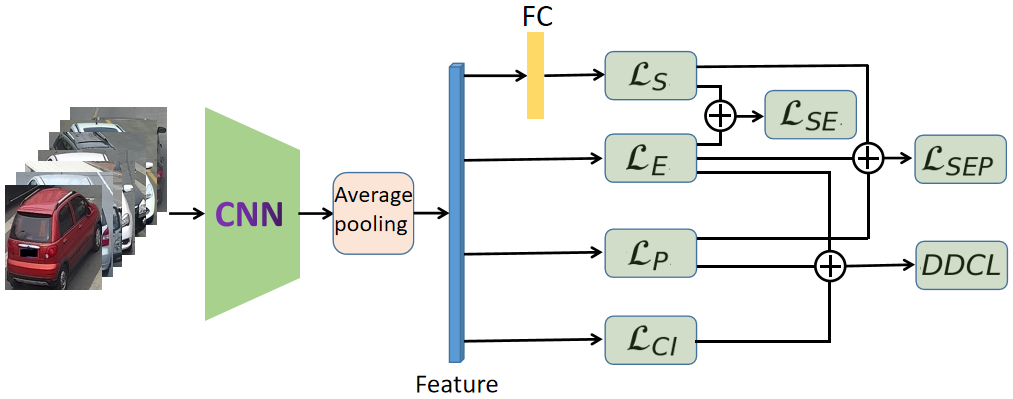}
	\caption{The structure used in this paper. The backbone network is resnet50. We change the output dimension of FC layer to the number of the training vehicle IDs. $\mathcal{L}_{SE}$ and the proposed $\mathcal{L}_{SEP}$ must combine with softmax loss, and the proposed DDCL has get rid of the limitation of softmax loss.}
	\label{network}       
\end{figure}

\subsection{The original center loss}
\label{sec:31}
\ \ \ \ The idea of center loss \cite{wen2016discriminative} is to add center parameters to the model and update these center parameters during training. The center loss is defined as follows:

\begin{equation}
	\label{LE}
	\mathcal{L}_E=\frac{1}{2m}\sum_{i=1}^{m}\Vert x_i-c_{y_i}\Vert _2^2
\end{equation}

\noindent where $m$ is the batch size, $x_i$ is the feature of the $i$-th image in the batch, $y_i$ is the ground truth of $x_i$ , and $c_{y_i}$  is the center of the $y_i$-th class, $\Vert \cdot \Vert _2$ is the $L_2$ norm. Finally, the following weighted loss function is used to supervise the training:

\begin{equation}
	\label{LSE}
	\mathcal{L}_{SE}=\mathcal{L}_{S}+\alpha\mathcal{L}_{E}
\end{equation}

\noindent where $\mathcal{L}_{S}$ is softmax loss, and $\alpha$  is the weight of  $\mathcal{L}_{E}$.

In the back-propagation, in order to prevent the large disturbance caused by a few mislabeled samples, the moving average strategy is used to control the update of the center parameters. The derivative of $\mathcal{L}_E$ to $x_i$  and $c_j$, and the update formula of center parameters are given as follows:

\begin{equation}
	\label{dledxi}
	\frac{\partial\mathcal{L}_E}{\partial x_i}=\frac{1}{m}( x_i-c_{y_i})
\end{equation}
\begin{equation}
	\label{deltacj}
	\Delta c_j=\frac{\sum_{i=1}^{m}\delta (y_i=j)\cdot (c_j-x_i)}{\xi + \sum_{i=1}^{m}\delta (y_i=j)}
\end{equation}
\begin{equation}
	\label{cjt1}
	c_j^{t+1}=c_j^t-\lambda \cdot \Delta c_j^t
\end{equation}
\noindent where $\delta (\cdot)$ is an indicator function, i.e. $\delta(True)$=1 and $\delta(False)$=0. $\xi$ is a very small positive number and $\lambda$ is the learning rate of the center parameters.

\subsection{Pearson distance center loss}
\label{sec:32}
\ \ \ \ Center loss \cite{wen2016discriminative} hopes to enhance the intra-class compactness by narrowing the Euclidean distance between features and their centers, but as the author of center loss \cite{wen2016discriminative} described in \cite{wen2019comprehensive}, we cannot overestimate the ability of center loss, because the attraction of center loss to some hard samples is not enough. In order to enhance the attraction of centers, we add a Pearson  center loss on formula \ref{LSE}, and use the following weighted loss function to supervise training:
\begin{equation}
	\label{lsep}
	\mathcal{L}_{SEP}=\mathcal{L}_{S}+\alpha\mathcal{L}_{E}+\beta\mathcal{L}_{P}
\end{equation}

\noindent where $\mathcal{L}_{P}$ is the Pearson center loss and $\beta$ is its weight. $\mathcal{L}_{P}$ is defined as follows:

\begin{equation}
	\label{LP}
	\mathcal{L}_{P}=\left[\frac{1}{m}\sum_{i=1}^{m}P(x_i,c_{y_i})\right]^\gamma=\left[1-\frac{1}{m}\sum_{i=1}^{m}C(x_i,c_{y_i})\right]^\gamma
\end{equation}

\noindent where $x_i=(x_{i,1},x_{i,2},\cdots,x_{i,n})^T$ is the feature of the $i$-th image in the batch, $c_{y_i}=(c_{y_i,1},c_{y_i,2},\cdots,c_{y_i,n})^T$ is the center of $x_i$, $y_i$ is the ground truth of $x_i$. $\gamma$ is a positive number and satisfies $\gamma >1$, which is used to control the stability of the whole model. $P(x_i,c_{y_i})$ is the Pearson distance between $x_i$ and $c_{y_i}$, and  $C(x_i,c_{y_i})$ is the Pearson correlation coefficient between $x_i$ and $c_{y_i}$, $C(x_i,c_{y_i})$ is defined as follows:
\begin{equation}
	\label{corr}
	C(x_i,c_{y_i})=\frac{\sigma _{x_i,c_{y_i}}}{\sigma _{x_i}\cdot \sigma _{c_{y_i}}}	=\frac{(x_i-\overline{x_i})^T(c_{y_i}-\overline{c_{y_i}})}{\Vert x_i-\overline{x_i}\Vert _2 \Vert c_{y_i}-\overline{c_{y_i}}\Vert _2}
\end{equation}

\noindent where $\sigma_{x_i}$ and $\sigma_{c_{y_i}}$ are the standard deviations of $x_i$ and $c_{y_i}$, respectively, $\sigma _{x_i,c_{y_i}}$ is the covariance between $x_i$ and $c_{y_i}$. $\overline{x}$ is a vector with the same size of  $x$ , and all of its elements are equal to the average of all components of $x$.

\begin{figure}
	\includegraphics[width=1\textwidth]{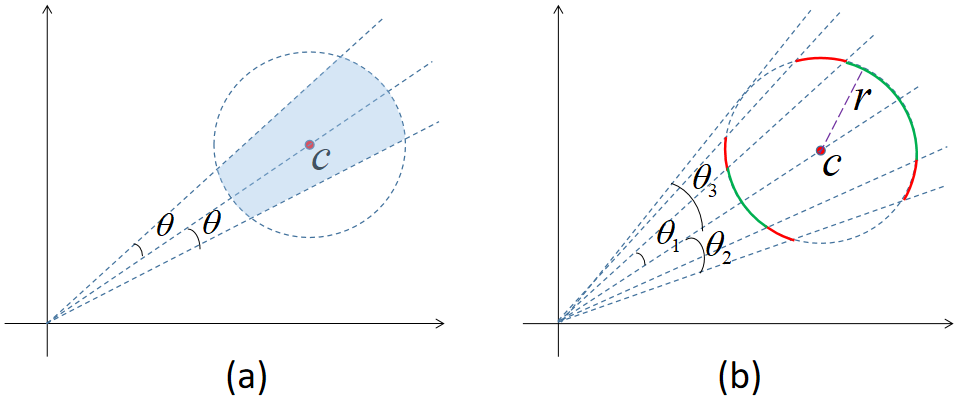}
	\caption{(a) the effect of $\mathcal{L}_E+\mathcal{L}_P$, (b) the effect of different $\gamma$. In(a) , the combination of $\mathcal{L}_E$ and $\mathcal{L}_P$ makes $x$ to be confined in the intersection of a hypersphere and a uncoated hypercone (the shaded part in the figure), making $x$ more closer to $c$. In (b), given $1<\gamma _1<\gamma_2$, when the distance between the feature and its center is fixed as r (i.e., the same iteration), the features of the model corresponding to $\gamma=\gamma_1$ are limited to the green arc, while the features of the model corresponding to $\gamma=\gamma_2$ are limited to the green arc and the red arc jointly (this does not affect the measurement of similarity, because we only use the Euclidean distance to calculate the similarity). The angle between the feature and its center cannot exceed $\theta_3$, so the model tends to be stable when $\gamma$ is greater than a certain value.}
	\label{intersection}       
\end{figure}

\noindent \textbf{Analysis:} Since Pearson distance (or cosine distance) and Euclidean distance are two completely different distances and have strong complementarity, combine  $\mathcal{L}_E$ and $\mathcal{L}_P$ can restrict the features of the same class into the intersection of a hypersphere and a hypercube in the feature space (see Fig. \ref{intersection} (a)), which will make the feature closer to its center. The value of $\gamma$ must be greater than 1, otherwise the $\frac{\partial \mathcal{L}_P}{\partial x_i}$ and $\frac{\partial \mathcal{L}_P}{\partial c_j}$ will be very large in the middle and late stages of training process, which will make $\mathcal{L}_{SEP}$ not converge. Because the test of the model is based on Euclidean distance, the angle does not affect the similarity.  For the same iteration of the training process (i.e. the same radius $r$ in Fig. \ref{intersection}), the greater the $\gamma$, the smaller the correlation between the feature and its center (see the appendix for proof), and the larger the angle between the feature and its center, which will provides more favorable conditions for training the hard samples. Since the angle between the feature and its center has an upper limit (which can not exceed $\theta _3$ in Fig. \ref{intersection} (b)), the model tends to be stable when $\gamma$ exceeds a certain value.

\subsection{Enhance the inter-class separability}
\label{sec:33}

\ \ \ \ The center loss \cite{wen2016discriminative} only takes into consideration of the intra-class compactness of features, but not the inter-class separability. Center loss \cite{wen2016discriminative,wen2019comprehensive}  described that $\mathcal{L}_E$ must be used together with $\mathcal{L}_S$. If $\mathcal{L}_E$ is used alone, all features and centers would be trained closed to 0. We believe that all features and centers will be trained closed to a same point in the feature space, but not 0, which can also make $\mathcal{L}_E$ closed to 0 when we minimize the formula \ref{LE}. This inspires us to focus on preventing all centers from being trained to a same point, rather than zero. So we design the following center isolation loss:

\begin{equation}
	\label{lci}
	\mathcal{L}_{CI}(d_{\mathcal{E}})=\frac{\sum\limits_{1\leq i <j\leq N}\delta(\Vert c_i-c_j\Vert _2^2<d_{\mathcal{E}})\cdot \Vert c_i-c_j\Vert _2^2}{\nu+\sum\limits_{1\leq i <j\leq N}\delta(\Vert c_i-c_j\Vert _2^2<d_{\mathcal{E}}) }
\end{equation}

\noindent where $N$ is the number of training vehicle IDs, $d_\mathcal{E}$ is the Euclidean distance threshold between all center pairs, $\nu$($>$0) is a small sample inhibition factor, with which, when the number of center pairs whose distance is less than $d_\mathcal{E}$ is too small,  can reduce the learning speed of the parameters of these center pairs and make the model more stable. So we remove $\mathcal{L}_S$ and use the following formula to train the model:

\begin{equation}
	\label{ddcl}
	\mathcal{L}_{EPCI}(d_{\mathcal{E}})=\alpha\mathcal{L}_E+\beta\mathcal{L}_P-\mu\mathcal{L}_{CI}(d_{\mathcal{E}})
\end{equation}
\noindent where $\mu$ is the weight of $\mathcal{L}_{CI}$, we call $\mathcal{L}_{EPCI}$ as dual distance center loss, or DDCL for short.

\noindent \textbf{Analysis: } $\mathcal{L}_{CI}$ prevents all centers from being trained to a same point. In this way,  $\mathcal{L}_{S}$ can be removed, the weight parameters and bias parameters of FC layer can be removed too (see Fig. \ref{network}). While the number of the weight parameters of FC layer is the same as that of center parameters, so we not only do not add more parameters, but reduce it. At the same time, the way we calculate  $\mathcal{L}_{CI}$ use the distance between all centers, which increases the participation of the center parameters. Therefor, the proposed  $\mathcal{L}_{CI}$ solves the last four shortcomings of original center loss.

\section{Experiment and analysis}
\label{sec:4}
\ \ \ \ In this section, we first introduce the datasets, evaluation protocol and training settings  used in the following experiments, then use experiments to verify that we have solved the five shortcomings of the original center loss, and conduct an in-depth analysis of $\mathcal{L}_{CI}$, we also discussed the impact of the small sample inhibition factor $\nu$ to the experimental accuracy, and then compared the experimental results of DDCL with that of center loss \cite{wen2016discriminative} in the four datasets. Finally, we compare the experimental results in the field of vehicle re-identification between DDCL and the state-of-the-art methods
\subsection{Experimental tools}
\subsubsection{Datasets}
\label{sec:411}
\ \ \ \ The veri-776 dataset is collected by Liu et al. \cite{liu2016large}. There are 576 vehicles in the training set, with total of 37778 images. The testing set contains 200 vehicles, which are divided into gallery set (11579 images in total) and query set (1678 images in total).  VehicleID dataset \cite{liu2016deep} is another widely used dataset in the field of vehicle re-identification. The training set of this dataset consists of 113346 images of 13164 vehicles. The test set is subdivided into three subsets: small, medium and large, which contain 6439, 13377 and 19777 images of  800, 1600 and 2400 vehicles, respectively. However, Liu et al. \cite{liu2016deep} did not give a fixed method of how to divide the three subsets into their respective query sets and gallery sets, so the unified approach is used to randomly select one image from each vehicle and put it into the gallery set and the rest into the query set. The training set of MSMT17 dataset contains 32621 images of 1041 persons, while the test set includes 93820 images of 3060 persons. The query set contains 11659 images, and the gallery set contains 82161 images. The training set of market1501 dataset contains 12936 images of 751 persons, while the test set contains 19732 images of 750 persons. The test set is divided into query set and gallery set, the query set contains 3368 images, and the gallery set contains 19732 images.

\subsubsection{Evaluation protocol}
\label{sec:411}

\textbf{mAP} Denote $P(q_i,k)$ and $R(q_i,k)$ represent the precision and recall rate of query image $q_i$ in the top-$k$ images, and assume $R(q_i,0)=0$, then AP (Average precision)and mAP (mean Average Precision) are defined as:
\begin{equation}
	\label{ap}
	AP(q_i)=\sum_{k=1}^{G}P(q_i,k)\delta \left[P(q_i,k)\neq P(q_i,k-1)\right]
\end{equation}
\begin{equation}
	\label{map}
	mAP=\frac{\sum\limits_{i=1}^{Q}AP(q_i)}{Q}
\end{equation}
\noindent where, $Q$ and $G$ represent the total number of images in query set and gallery set, respectively, for $\delta(\cdot)$, there are $\delta(True)=1$ and $\delta(False)=0$.

\noindent \textbf{CMC} CMC (Cumulative Matching Characteristics) is defined as:
\begin{equation}
	\label{cmc}
	CMC@k=\frac{\sum\limits_{i=1}^{Q}h(q_i,k)}{Q}
\end{equation}
\noindent where $Q$ is the total number of images in query set, and $q_i$ is the $i$-th image in query set. When using $q_i$ to query in the gallery set, if there is a correct query in top-$k$, then $h(q_i,k)=1$, otherwise $h(q_i,k)=0$.

\subsubsection{Training configurations}
\ \ \ \ \ We use resnet50 as our backbone network. For all the convolutional part, we use the pretrained model on ImageNet to initialize the network. The parameter initialization methods of the FC layer (see Fig. \ref{network}) will be discussed later. For res4 (the last convolutional block) of resnet50, the stride is set to 1. In the training stage and testing stage, the input images are all processed as follows: (1) resize image size to $224\times 224$, (2) the pixel value is linearly adjusted to [0,1], (3) the mean and standard deviation are set to [0.485, 0.456, 0.406] and  [0.229, 0.224, 0.225] as the same as  the ImageNet. Using Adam parameter optimization strategy, the batch size is set to 64. As in \cite{wen2016discriminative}, we set $\lambda$ to 0.5, $\alpha$ to 0.003, and we set $\mu$ in formula \ref{ddcl} to 0.005. The learning rate begins at 0.0001. The learning rate adjustment strategy is as follows. For the VeRi-776 dataset, MSM T17 dataset and market1501 dataset, the learning rate is divided by 10 at 10 and 17 epoch, and the total epochs is 22. For the VehicleID dataset, the learning rate is divided by 10 at 50 and 60 epoch, and the total epochs is 65. 

\subsection{The analysis of generalization ability of the original center loss}
\label{sec:42}
\begin{figure}
	\includegraphics[width=1\textwidth]{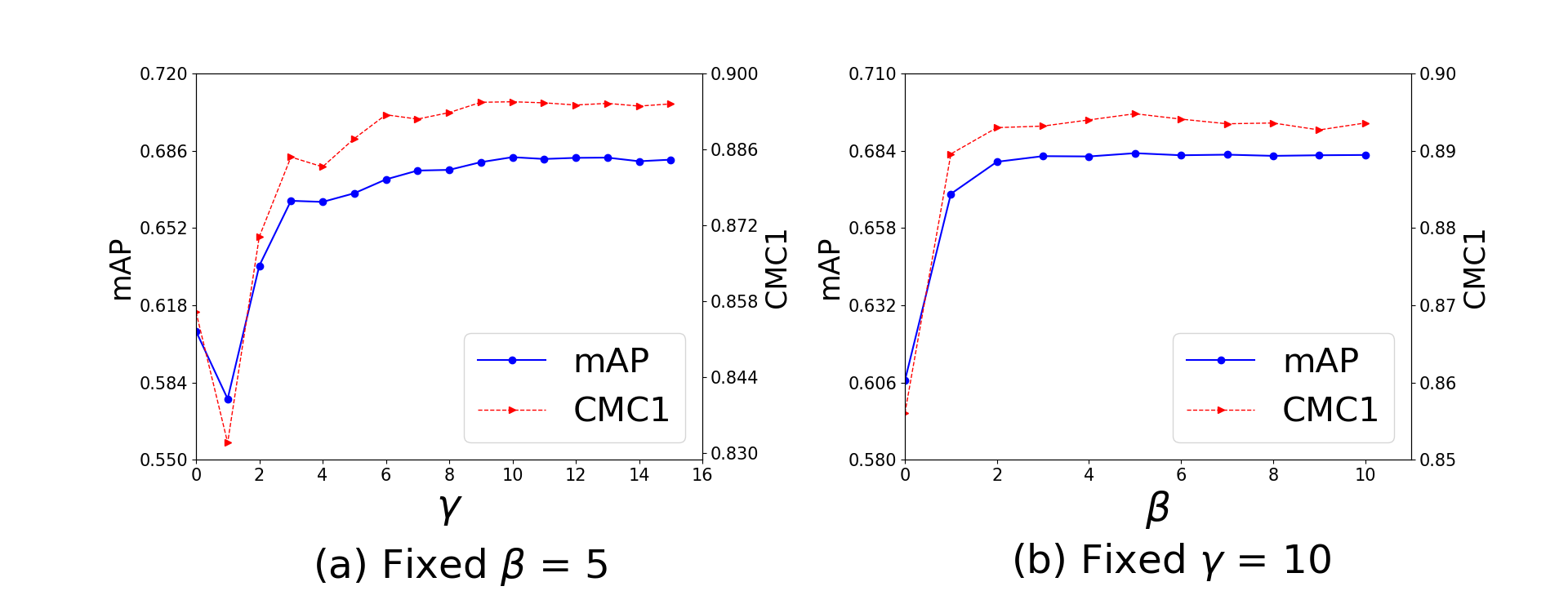}
	\caption{Fixed $\beta=5$ and $\gamma=10$ respectively. With the change of $\gamma$ and $\beta$  the trend of mAP and CMC1.}
	\label{beta_gamma}       
\end{figure}

\textbf{The value of $\beta$ and $\gamma$}  Before discussing the generalization ability , it is necessary to use experiment to determine the values of $\beta$ and $\gamma$ in formulas \ref{lsep} and \ref{LP}. In this experiment, we use the Normal distribution with std=0.001 and mean=0 to initialize the weight parameters of the FC layer, and the bias parameters are initialized as 0. We first fix $\beta=5$ to find the optimal $\gamma$, and then fix $\gamma=10$ to find the optimal $\beta$. The experimental results are shown in Fig. \ref{beta_gamma}.

In Fig. \ref{beta_gamma}, both $\beta=0$ and $\gamma=0$ represent the original center loss \cite{wen2016discriminative} (i.e. $\mathcal{L}_{SE}$). As can be seen from Fig. \ref{beta_gamma} (a), the mAP and CMC1 of $\mathcal{L}_{SEP}$ (formula \ref{lsep}) is much higher than that of $\mathcal{L}_{SE}$ (see $\beta=0$) for all $\gamma$ except $\gamma=1$. When $\gamma$ changes in 1,2 and 3, the mAP and CMC1 change greatly, which indicates that $\gamma$ has great influence on $\mathcal{L}_{SEP}$. When $\gamma$ is bigger than 5, the mAP and CMC1 change smoothly , which verifies the conclusion in section \ref{sec:32} that the model tends to be stable when $\gamma$ is greater than a certain value.  Fig. \ref{beta_gamma} (b) shows that, when fixed $\gamma=10$, $\beta$  has little influence on the mAP and CMC1 of our proposed method. According to Fig. \ref{beta_gamma}, we set $\beta$ to 5 and $\gamma$ to 10 in all of the following experiments.

\begin{figure}
	\includegraphics[width=1\textwidth]{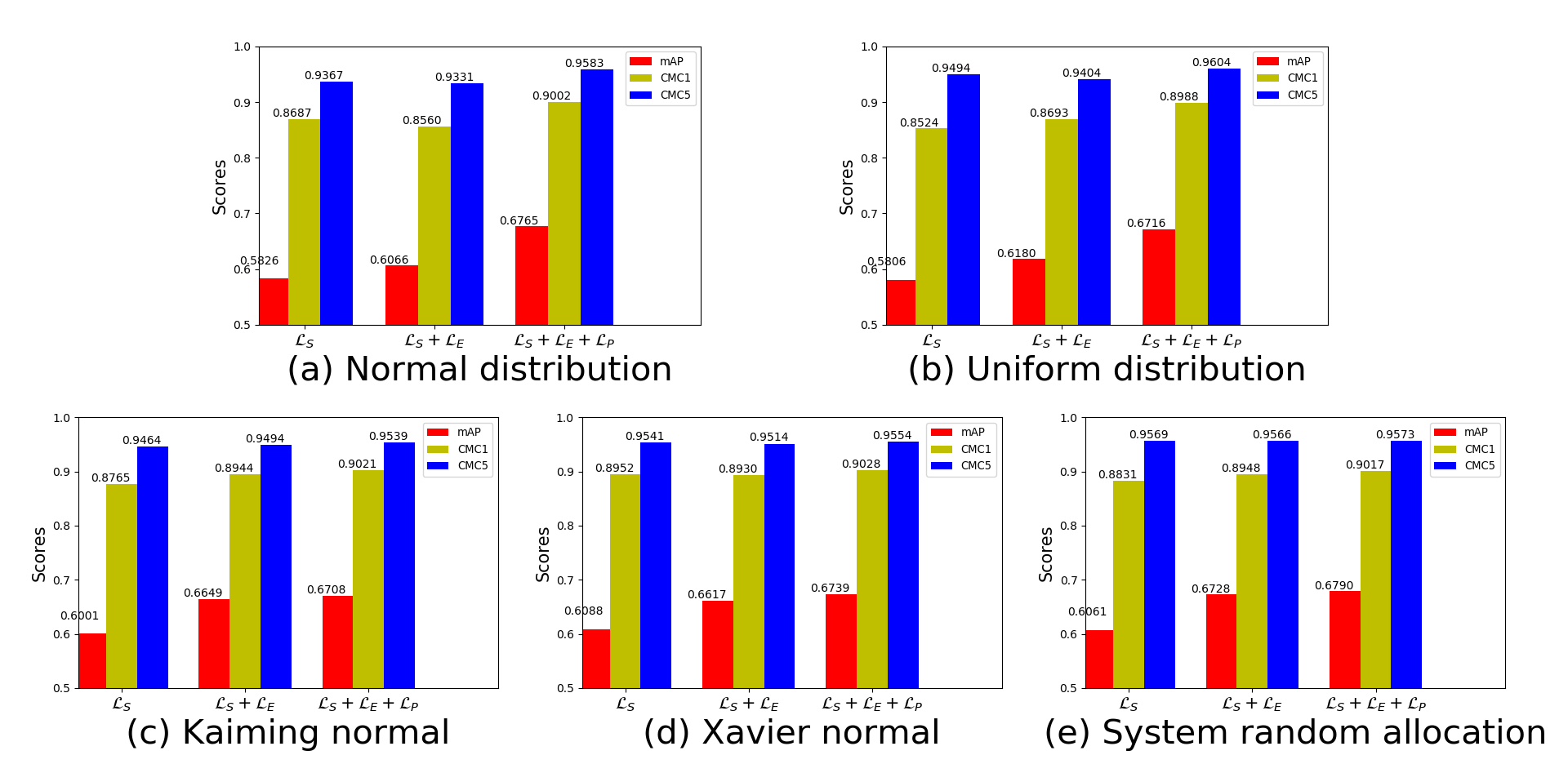}
	\caption{Five examples of different initialization methods for the weight parameters of the FC layer.}
	\label{generalization}       
\end{figure}

\noindent \textbf{Generalization ability analysis} $\mathcal{L}_{SE}$ is sensitive to the initialization methods for the weight parameters of the FC layer, because it performs well for some initialization methods, but not so well for others. Let's look at some examples of different initialization methods for the weight of the FC layer. The bias parameters of the FC layers are initialized as 0 for all the following methods, and the weight parameters are initialized as follows: (1) Normal distribution with std=0.001 and mean=0, (2) Uniform distribution with interval of [0,0.0001], (3) Kaiming normal, (4) Xavier normal, (5) system random allocation. The experimental results are shown in Fig. \ref{generalization}. The Fig. \ref{generalization} shows that for the Normal distribution initialization method, the mAP of $\mathcal{L}_{S}$ + $\mathcal{L}_{E}$ (i.e. $\mathcal{L}_{SE}$) is 2.4\% higher than that of $\mathcal{L}_{S}$, but the CMC1 and CMC5 are 1.27\% and 0.46\% lower than that of $\mathcal{L}_{S}$, respectively. After adding $\mathcal{L}_{P}$  (i.e. $\mathcal{L}_{S}$+$\mathcal{L}_{E}$+$\mathcal{L}_{p}$ or $\mathcal{L}_{SEP}$), compared with $\mathcal{L}_{S}$+$\mathcal{L}_{E}$, mAP, CMC1 and CMC5 are improved by 6.99\%, 4.42\% and 2.52\%, respectively. For Uniform distribution initialization method, mAP and CMC1 of $\mathcal{L}_{S}$+$\mathcal{L}_{E}$ are 3.24\% and 1.69\% higher than that of $\mathcal{L}_{S}$, while CMC5 reduces by 0.9\%. After adding $\mathcal{L}_{p}$, mAP, CMC1 and CMC5 of $\mathcal{L}_{S}$+$\mathcal{L}_{E}$+$\mathcal{L}_{P}$ are 5.36\%, 2.95\% and 2\% higher than that of $\mathcal{L}_{S}$+$\mathcal{L}_{E}$, respectively, which is also a great improvement. For the later three initialization methods (i.e. (3) (4) (5)), $\mathcal{L}_{S}$+$\mathcal{L}_{E}$ has been greatly improved than $\mathcal{L}_{S}$, which seems that the role of $\mathcal{L}_{p}$ is weakened, but after adding $\mathcal{L}_{p}$,  the mAP, CMC1 and CMC5 promote to some extent. The two initialization methods of (1) (2) are examples of insufficient generalization ability of $\mathcal{L}_{S}$+$\mathcal{L}_{E}$ \cite{wen2016discriminative}. From the five subgraphs in Fig. \ref{generalization}, we find that mAP, CMC1 and CMC5 of $\mathcal{L}_{S}$+$\mathcal{L}_{E}$+$\mathcal{L}_{p}$ are almost the same. Therefore, $\mathcal{L}_{p}$ increases the generalization ability of the center loss.

\begin{figure}
	\includegraphics[width=1\textwidth]{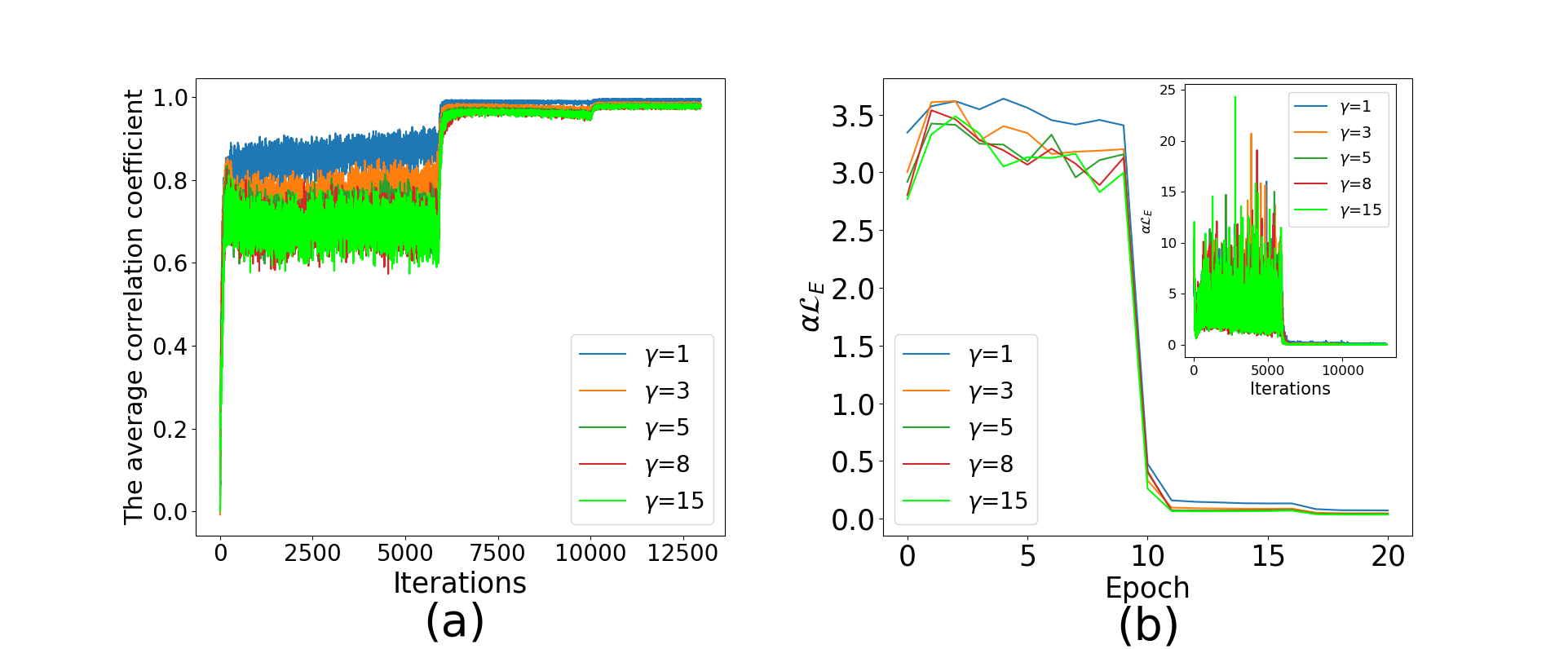}
	\caption{The effect of $\gamma$. The big graph in (b) is the result of averaging the $\alpha\mathcal{L}_E$ $(\alpha=0.003)$ in the subgraph in an epoch. In (a), the larger $\gamma$, the smaller the average correlation coefficient between the features and their centers in a batch. In (b), the larger $\gamma$, the smaller the Euclidean distance between the features and their centers.}
	\label{corr_eu}       
\end{figure}

In order to observe the effect of $\gamma$ on $\mathcal{L}_{SEP}$, we draw Fig. \ref{corr_eu}. The subgraph in Fig. \ref{corr_eu}(b) represents the changes of $\alpha\mathcal{L}_E$ under different $\gamma$. We know from formula \ref{LE} that $\mathcal{L}_E$  is the average Euclidean distance between the features and their centers in the batch. Because $\mathcal{L}_E$ fluctuates too heavily, we reassign the values of $\alpha\mathcal{L}_E$ greater than 7 to 7, and then average them in each epoch to obtain the big graph in Fig. \ref{corr_eu}(b). In Fig. \ref{corr_eu}, it can be seen that for the same $\gamma$, as iterations increase, the average correlation coefficient increases (Fig. \ref{corr_eu}(a)), and the Euclidean distance decreases ( Fig. \ref{corr_eu}(b)). For the same moment of training (i.e. the same iteration), the larger the $\gamma$, the smaller the Euclidean distance (Fig. \ref{corr_eu}(b)), and the smaller the average correlation coefficient (Fig. \ref{corr_eu}(a)), which is the same as the analysis in section \ref{sec:32}. Since the test is based on the Euclidean distance to measure the similarity, the smaller the Euclidean distance, the better the model, indicating that a larger $\gamma$ will have a better effect. Of course, it’s no need to set $\gamma$ to a very large value, when $\gamma$ is too large, the model has become stable. The average correlation coefficient curves of $\gamma=$5, 8  and 15 in Fig. \ref{corr_eu}(a) almost overlap, and the Euclidean distance curves of $\gamma=$5, 8  and 15  in Fig. \ref{corr_eu}(b) almost overlap too, indicating that when $\gamma$ is greater than 5, the model has became stable, which is the same conclusion as the analysis in Fig. \ref{beta_gamma} and section \ref{sec:31}.

\subsection{Enhance the inter-class separability}
\ \ \ \ In this section, we discuss the following three shortcomings of the center loss \cite{wen2016discriminative}, (1) it does not take the consideration of the inter-class separability; (2) it must be used with the combination of $\mathcal{L}_S$; (3) it needs to add additional parameters. In fact, the solution to the three shortcomings is essentially the same, because we prevent all centers from being trained to the same point, the inter-class separability is considered, and $\mathcal{L}_S$ can be removed, then the FC layer is removed accordingly (see Fig. \ref{network}), the total number of the model parameters does not increase. So the three shortcomings are improved simultaneously through the proposed $\mathcal{L}_{CI}$.

\begin{figure}
	\includegraphics[width=1\textwidth]{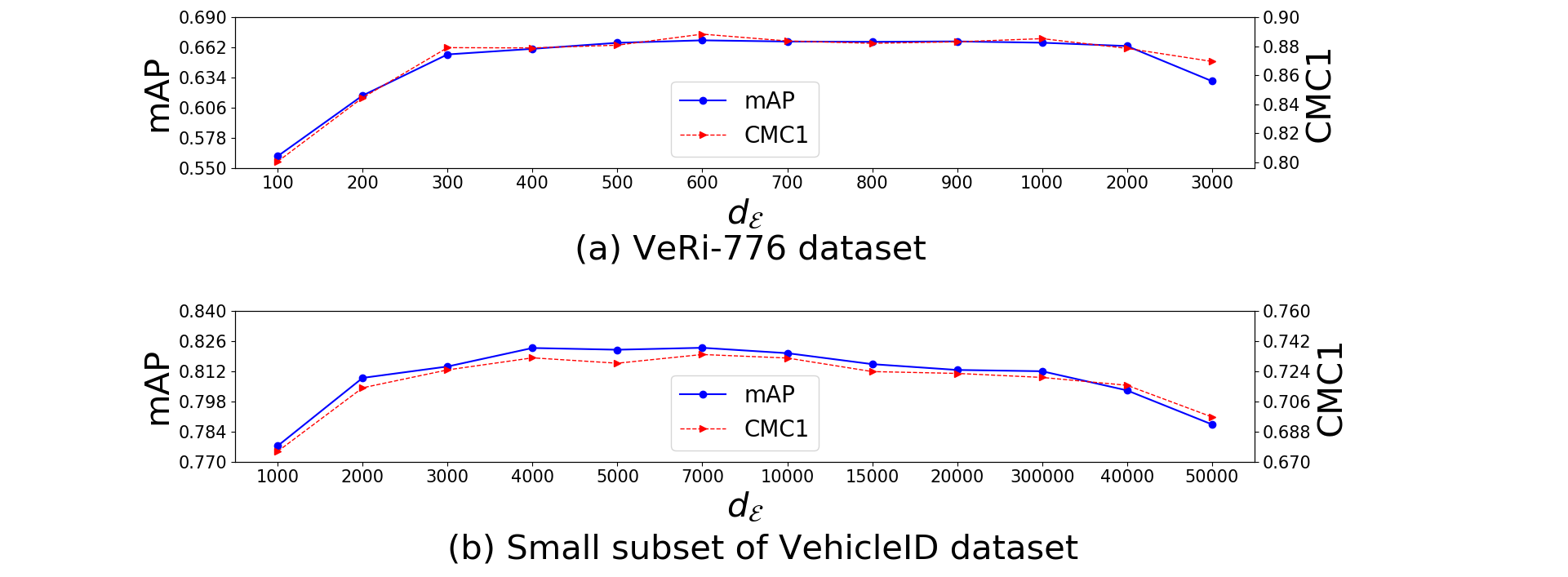}
	\caption{In the VeRi-776 dataset and the small subset of VehicleID dataset, the effect of different $d_\mathcal{E}$ on DDCL.}
	\label{de}       
\end{figure}

Next, we use experiments to verify that DDCL can train models without the combination of $\mathcal{L}_E$. Fig. \ref{de} shows the effect of different $\mathcal{L}_E$ on DDCL in the VeRi-776 dataset and the small subset of VehicleID dataset. As can be seen from Fig. \ref{de}, due to the small sample inhibition factor $\nu$ designed in formula \ref{lci}, the experimental results with $d_\mathcal{E}$ from 300 to 2000 in the VeRi-776 dataset is almost the same, with  $d_\mathcal{E}$ from 3000-10000 in the small subset of VehicleID dataset is almost the same, and in the small subset of VehicleID dataset when  $d_\mathcal{E}$ is bigger than 10000 the experimental results decrease slowly. All of the above proves the strong adaptability of DDCL.

\begin{figure}
	\includegraphics[width=1\textwidth]{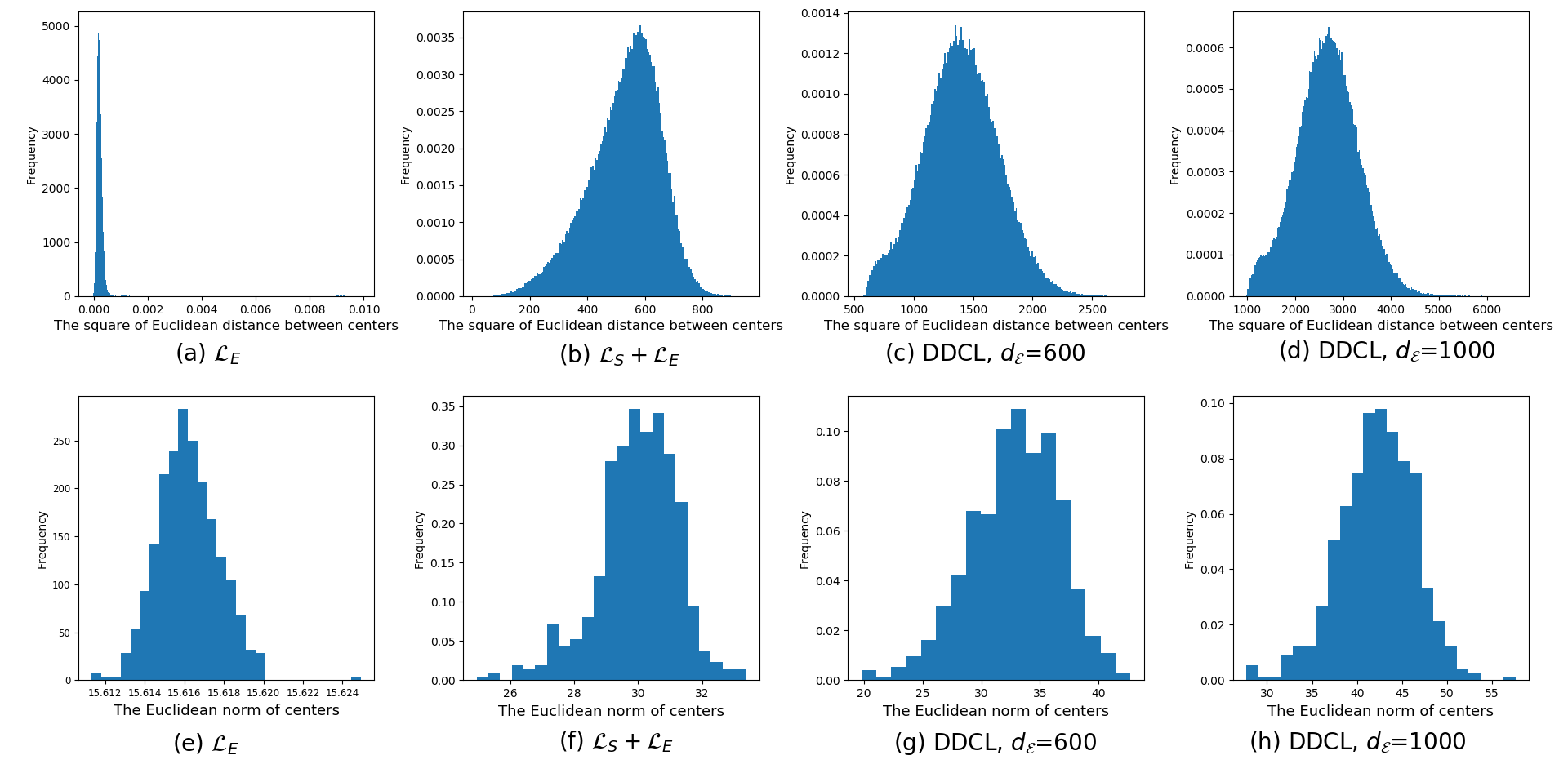}
	\caption{Under the supervision of different loss functions, the square of the Euclidean distance between centers and the Euclidean norm of centers in the VeRi-776 dataset.}
	\label{center_dist_norm}       
\end{figure}
In order to further observe the working mechanism of $\mathcal{L}_{CI}$, in the VeRi-776 dataset, we draw the histogram of the square of the distance between centers and the histogram of Euclidean norm of centers, these centers are obtained by training with the supervision of the following loss functions, (1) $\mathcal{L}_{E}$ \cite{wen2016discriminative}; (2) $\mathcal{L}_{SE}$ \cite{wen2016discriminative}; (3)DDCL $(d_\mathcal{E}=600)$; (4)DDCL $(d_\mathcal{E}=1000)$. It is observed that in Fig. \ref{center_dist_norm}(a) the distances between the center pairs obtained by $\mathcal{L}_{E}$ are very close to 0  and Fi. \ref{center_dist_norm}(b) shows that the Euclidean norms of all centers obtained by $\mathcal{L}_{E}$ are very close to 15.616, indicating that $\mathcal{L}_{E}$ makes all the centers close to a same point, not zero, which verifies the conclusion in section \ref{sec:33}. Fig. \ref{center_dist_norm}(b)(f) show that $\mathcal{L}_{SE}$ makes all centers close to a hypersphere with a radius of 30, (c)(d)(g)(h) show that the radius of the hypersphere formed by the centers trained by DDCL is larger than that formed by the centers trained by $\mathcal{L}_{SE}$, which indicates that $\mathcal{L}_{S}$ and DDCL are inconsistent in the feature space. It can also be seen from Fig. \ref{center_dist_norm} that the larger the $d_\mathcal{E}$, the larger the radius of the hypersphere formed by the centers obtained by DDCL.

\begin{figure}
	\centering\includegraphics[width=.6\textwidth]{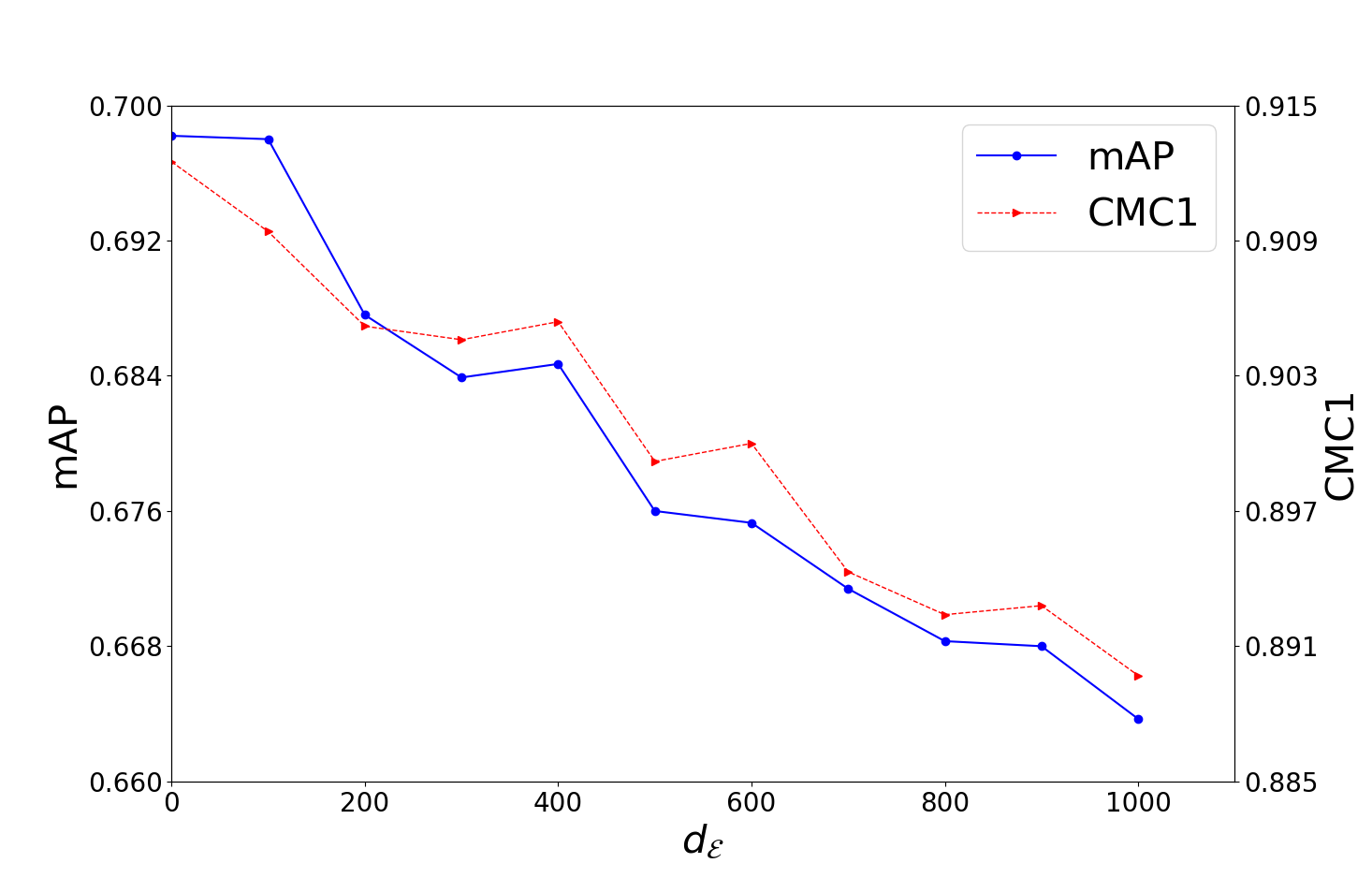}
	\caption{In the veri-776 dataset,  combine $\mathcal{L}_S$ and DDCL to train the model. With the increase of $d_\mathcal{E}$, the mAP and CMC1 decrease.}
	\label{inconsistent}       
\end{figure}

It is mentioned above that DDCL and $\mathcal{L}_S$ are inconsistent in the feature space. In order to verify this conclusion, we do the following experiments in the VeRi-776 dataset, we use  DDCL combines with $\mathcal{L}_S$ to jointly train the model. It can be seen from Fig. \ref{inconsistent} that the experimental accuracy decreases with the increase of $d_\mathcal{E}$, and the experimental accuracy never exceeds the experimental accuracy of $\mathcal{L}_{SEP}$ (i.e., $d_\mathcal{E}$ = 0 in Fig. \ref{inconsistent}), indicating that the larger the  $d_\mathcal{E}$, the more incompatible between $\mathcal{L}_S$ and DDCL. The existence of $\mathcal{L}_{CI}$ influences the feature space of $\mathcal{L}_S$, so we believe that $\mathcal{L}_S$ and DDCL are inconsistent in feature space.

\begin{figure}
	\centering\includegraphics[width=1\textwidth]{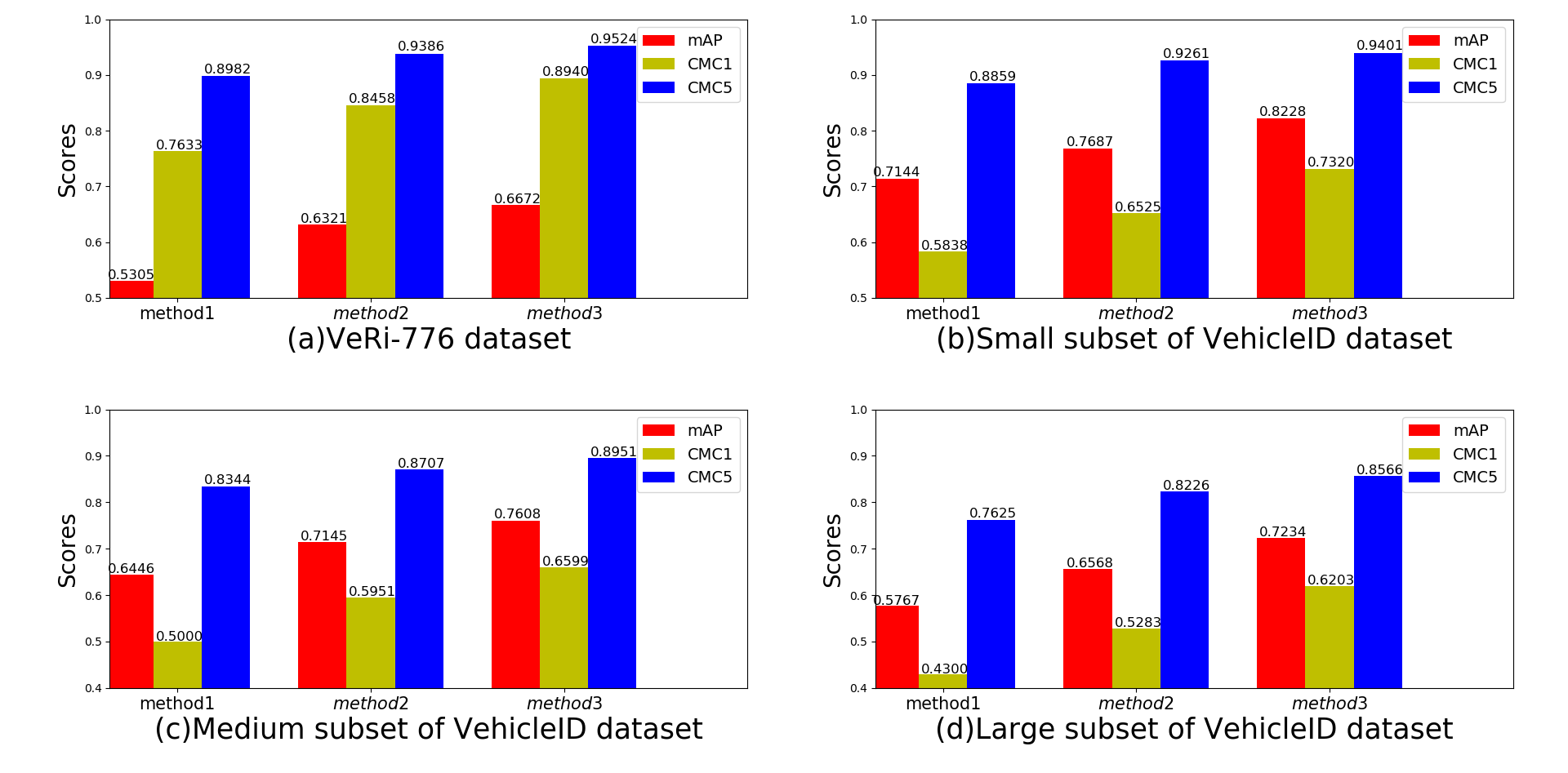}
	\caption{The influence of center parameter participation on experimental results in VeRi-776 dataset and VehicleID dataset}
	\label{center_participation}       
\end{figure}

\subsection{The analysis of Center Participation}

\ \ \ \ \ In order to prove that by increasing the participation of center parameters can improve the performance of DDCL, when calculating $\mathcal{L}_{CI}$, we compare the following three methods. (1) Only calculate the distance between the centers of the features which are contained in the batch (method1), ( 2) Calculate the distance between the centers of the features which are contained in the batch and all global centers (method2), (3) Calculate the distance between all global centers (method3). Obviously, the participation of center parameters of these three methods is sequentially increasing. Fig. \ref{center_participation} shows the experimental results of the three methods.

As can be seen from Fig. \ref{center_participation}, in the VeRi-776 dataset, mAP, CMC1 and CMC5 of method2 are 10.16\%, 8.25\% and 3.79\% higher than that of method1, respectively, and mAP, CMC1 and CMC5 of method3 are 3.41\%, 4.82\% and 1.38\% higher than that of method2, respectively. And in the three subsets of VehicleID dataset, we can get the same conclusions of VeRi-776 dataset. With the increase of the  participation of the center parameters, the experimental accuracy increases significantly, indicating that center participation has a great influence on DDCL. But we only increase the participation of center parameters in the aspect of inter-class separability, we can't find a way to increase its participation in the aspect of intra-class compactness, because this is limited by the number of images of each vehicle in the database, which is the insurmountable shortcoming of center loss. For the two loss functions of contrastive loss and triplet loss, and for each input, whether in the aspect of intra-class compactness or inter-class separability, all the parameters (including the parameters of the convolutional layer and the the parameters of the fully connected layer) participate in the optimization. But for the center loss, in the optimization of the aspect of intra-class compactness, one feature can only be used to optimize the center parameters belongs to its own center, and can not be used to optimize the parameters of other centers.

\subsection{The analysis of the small sample inhibition factor $\nu$}
\label{sec:45}
\label{sec:44}

\begin{figure}
	\centering\includegraphics[width=.75\textwidth]{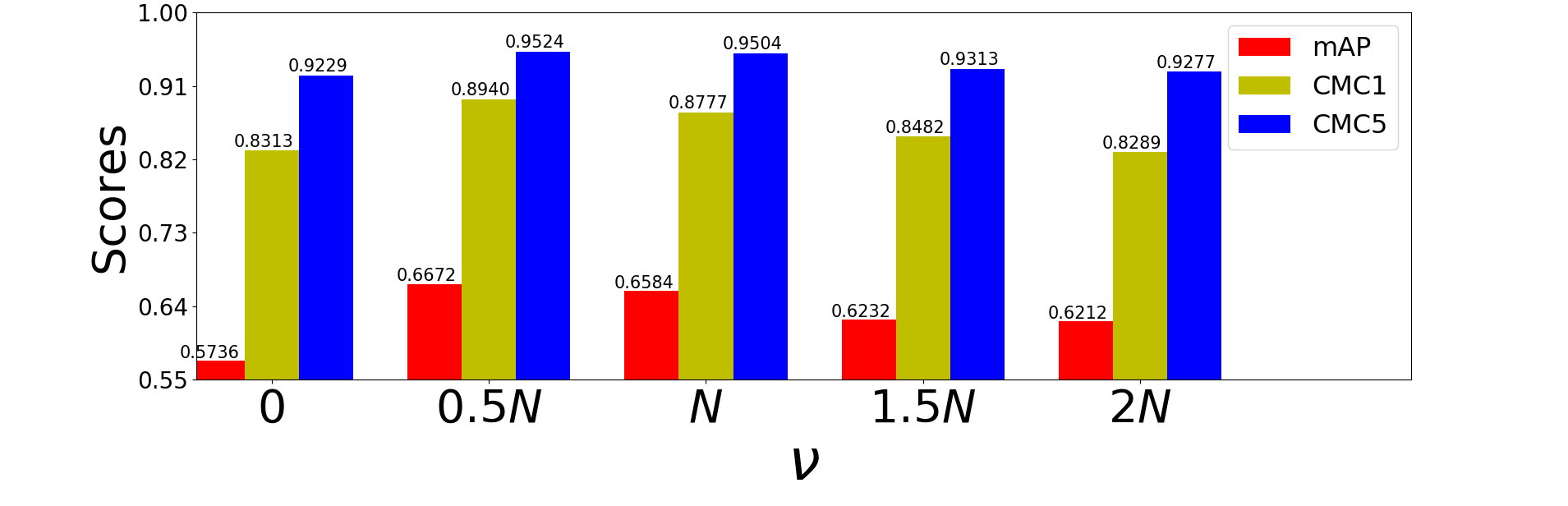}
	\caption{The effect of $\nu$ on the experimental results in the VeRi-776 dataset ($N$ denotes the number of classes in the training set)}
	\label{nu}       
\end{figure}

\ \ \ \ In order to verify the effect of the parameter $\nu$ designed in formula \ref{lci}, we conducte an experimental analysis on different $\nu$ in the VeRi-776 dataset. As shown in Fig. \ref{nu}, it is obsvered that $\nu$ has an great influence on the experimental results. For example, mAP, CMC1 and CMC5 of $\nu$ equals to 0.5$N$ are  9.36\%, 6.27\% and 2.95\% higher than that of $\nu$ equals to 0, respectively, where $N$ is the number of classes of the training set. So a good $\nu$ can greatly improve the experimental accuracy. According to Fig. \ref{nu}, we set $\nu$ to 0.5$N$.

\subsection{Comparison with original center loss}
\label{sec:46}

\begin{figure}
	
	\includegraphics[width=1\textwidth]{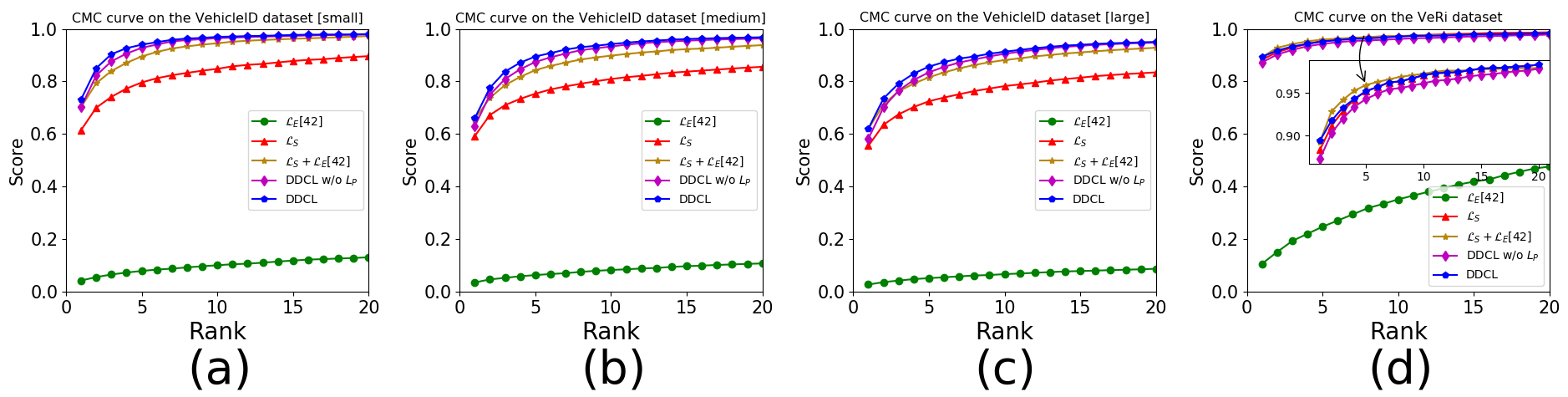}
	\caption{The CMC curves in the VeRi-776 dataset and VehicleID dataset.}
	\label{cmc_curve}       
\end{figure}

\ \ \ We conduct the two experiments on the center loss \cite{wen2016discriminative} by using the two following loss functions, (1) $\mathcal{L}_S+\mathcal{L}_E$ (i.e. $\mathcal{L}_{SE}$), (2) $\mathcal{L}_E$ (i.e. without $\mathcal{L}_S$), and compare the results with our proposed DDCL. We also conduct an experiment to remove $\mathcal{L}_P$ from DDCL (i.e. use $\mathcal{L}_E+\mathcal{L}_{CI}$)to verify the effect of $\mathcal{L}_P$. At the same time, we also list the experimental results of $\mathcal{L}_S$. We do the above experiments in two widely used datasets in the field of vehicle re-identification, named VeRi-776 dataset and VehicleID dataset, and in two widely used datasets in the field of person re-identification, named Market1501 and MSMT17. The experimental results are shown in Table \ref{tab1}, Table \ref{tab2}, Table \ref{tab3} and Table \ref{tab4}. Then we plot the CMC curve about the experimental results in VehicleID dataset and VeRi-776 dataset in Fig. \ref{cmc_curve}.

\noindent \textbf{DDCL compares with $\mathcal{L}_E$} It can be observed from Table \ref{tab1}, Table \ref{tab2}, Table \ref{tab3} and Table \ref{tab4} that, in the four datasets,  all the experimental results of only using the $\mathcal{L}_E$ to supervise the CNN are very poor, they are all the same as the validating results of such a model, the parameters of the model are randomly assigned by the system and without any training, indicating that the $\mathcal{L}_E$ \cite{wen2016discriminative} can not be used for supervising training without $\mathcal{L}_S$. While all the experimental results of our proposed DDCL achieve a very high level, which far higher than only use $\mathcal{L}_E$, indicating our DDCL indeed can training model independently.

\begin{table}[h]
	\caption{The comparison between DDCL and center loss \cite{wen2016discriminative} in VehicleID dataset.}\label{tab1}
	\centering
	\setlength{\tabcolsep}{0.2mm}
	\begin{tabular}{cccccccccc}
		\hline
		\multirow{2}*{Method} &  \multicolumn{2}{c}{Small} &  \multicolumn{2}{c}{Medium} & \multicolumn{2}{c}{Large}\\
		\cline{2-10}
		~& mAP & CMC1 & CMC5 & mAP & CMC1 & CMC5 & mAP & CMC1 & CMC5 \\
		\hline
		
				$\mathcal{L}_E $ &0.0651 &0.0419 &0.0787 &0.0532&0.0352&0.0629 & 0.0424 & 0.0272& 0.0514\\
		$\mathcal{L}_S$ & 0.6957 &0.6164 &0.7970 &0.6679&0.5921	&0.7534 &0.6338 &0.5551& 0.7246\\
				$\mathcal{L}_{SE} $ &0.7865	 &0.7035 &0.8958 &0.7380&0.6565	&0.8423	 &0.7085 &0.6191 &0.8158 \\
		\tabincell{c}{DDCL \\  $d_\mathcal{E}=4000$ \\ w/o $\mathcal{L}_P$ (ours)}&0.8001&0.7036&0.9281 &0.7362&0.6290&0.8742&0.6985&0.5819 &0.8349 \\
		\tabincell{c}{DDCL \\ $d_\mathcal{E}=4000$ (ours)} & \textbf{0.8228} & \textbf{0.7320} &\textbf{0.9401}	&\textbf{0.7608}&\textbf{0.6599}&\textbf{0.8951} &\textbf{0.7234}&\textbf{0.6203}   &\textbf{0.8566} \\

		\hline

	\end{tabular}
\end{table}

\begin{table}[h]
	\caption{The comparison between DDCL and center loss  in VeRi-776 dataset.}
	\label{tab2}
	\centering
	\setlength{\tabcolsep}{6mm}
	\begin{tabular}{cccc}
		\hline
		Method & mAP & CMC1 & CMC5 \\
		\hline
				$\mathcal{L}_E $ &0.0262&0.1060	&0.2470	 \\
		$\mathcal{L}_S$ &0.6061	&0.8831	&\textbf{0.9569} \\
		$\mathcal{L}_{SE}$ &\textbf{0.6728}	&\textbf{0.8948}&0.9566 \\
		DDCL $d_\mathcal{E}=600$ w/o $\mathcal{L}_P$ (ours) &0.6223	&0.8512	&0.9337\\
		DDCL $d_\mathcal{E}=600$ (ours) &0.6672	&0.8940	&0.9524	 \\		
		
		\hline	
	\end{tabular}
\end{table}

\begin{table}[h]
	\caption{The comparison between DDCL and center loss  in MSMT17 dataset.}
	\label{tab3}
	\centering
	\setlength{\tabcolsep}{6mm}
	\begin{tabular}{cccc}
		\hline
		Method &  mAP & CMC1 & CMC5 \\
		\hline
				$\mathcal{L}_E $ &0.0013&0.0042	&0.0127 \\
		$\mathcal{L}_S$ &0.2368	&0.4570	&0.6307	 \\
		$\mathcal{L}_{SE}$ &0.2865	&0.5114	&0.6929	 \\
		DDCL $d_\mathcal{E}=600$ w/o $\mathcal{L}_P$ (ours) &0.2698	&0.4826	&0.6658	 \\
		DDCL $d_\mathcal{E}=600$ (ours) &\textbf{0.3330}&\textbf{0.5609}&\textbf{0.7270} \\		
		
		\hline	
	\end{tabular}
\end{table}

\begin{table}[h]
	\caption{The comparison between DDCL and center loss  in Market1501 dataset.}
	\label{tab4}
	\centering
	\setlength{\tabcolsep}{6mm}
	\begin{tabular}{cccc}
		\hline
		Method & mAP & CMC1 & CMC5 \\
		\hline
				$\mathcal{L}_E $ &0.0109&0.0324	&0.0908	 \\
		$\mathcal{L}_S$ &0.5049	&0.6943	&0.8692	 \\
		$\mathcal{L}_{SE}$ &0.5693	&\textbf{0.7354}&\textbf{0.8945} \\
		DDCL $d_\mathcal{E}=600$ w/o $\mathcal{L}_P$ (ours) &0.5119	&0.6640	&0.8506	 \\
		DDCL $d_\mathcal{E}=600$ (ours) &\textbf{0.5730}&0.7262	&0.8930 \\		
		
		\hline	
	\end{tabular}
\end{table}

\noindent \textbf{DDCL compares with $\mathcal{L}_S$} Softmax loss ($\mathcal{L}_S$) neither takes the consideration of intra-class compactness nor inter-class separability, it only concern about the classification ability of the whole model. While DDCL not only takes consideration of intra-class compactness, but also the inter-class separability. From Table \ref{tab1}, Table \ref{tab2}, Table \ref{tab3} and Table \ref{tab4}, we can observe that, in the VeRi-776 dataset, CMC5 of DDCL is 0.45\% lower than that of $\mathcal{L}_S$, but mAP and CMC1 of DDCL is 6.11\% and 1.09\% higher than that of $\mathcal{L}_S$, we believe that our DDCL has exceeded $\mathcal{L}_S$ on this dataset. The reason for this is that the performance of DDCL is depend on the number of training IDs (this will be explained in the next section), the number of training IDs of VeRi-776 dataset in the four datasets is the smallest.  In other three datasets, all the mAPs, CMC1s and CMC5s of DDCL far exceed that of $\mathcal{L}_S$, especially in the VehicleID dataset, the reason is that the the number of training IDs of VehicleID dataset is the largest in the four datasets.

\noindent \textbf{DDCL compares with $\mathcal{L}_{SE}$} Let us list out the number of training IDs of the four datasets from small to large, VeRi-776 (576), Market1501 (751), MSMT17 (1041), VehicleID (13164), the numbers in the brackets are the number of training IDs in corresponding dataset. It can be observed that in the VeRi-776 dataset, mAP, CMC1 and CMC5 of DDCL are all lower than that of $\mathcal{L}_{SE}$, in Market1501 dataset, mAP of DDCL is higher than that of $\mathcal{L}_{SE}$, CMC1 and CMC5 of DDCL are lower than that of $\mathcal{L}_{SE}$. And in the rest two datasets, all the mAPs, CMC1s and CMC5s of DDCL are much higher than that of $\mathcal{L}_{SE}$. So, we can conclude that the larger the number of training IDs of a dataset, the better the performance of DDCL. This is because, in the formula \ref{lci}, we compute the $\mathcal{L}_{CI}$ by using the distance between all global centers, which makes more participation for the center parameters in the dataset with larger number of training IDs.

\noindent \textbf{DDCL compares with DDCL (without $\mathcal{L}_P$)} $\mathcal{L}_{CI}$ makes DDCL run without $\mathcal{L}_S$, and also can make $\mathcal{L}_E$ run without $\mathcal{L}_S$, i.e. DDCL (without $\mathcal{L}_P$). It can be seen from Table \ref{tab1}, Table \ref{tab2}, Table \ref{tab3} and Table \ref{tab4}, the mAP of DDCL about 2\%-7\% higher than that of DDCL (without $\mathcal{L}_P$) in all the four datasets, CMC1 of DDCL is about 3\%-7.5\% higher than that of DDCL (without $\mathcal{L}_P$), and CMC5 of DDCL is 1.2\%-6\% higher than that of DDCL (without $\mathcal{L}_P$), indicating the proposed $\mathcal{L}_P$ is very effective.

As can be seen in Fig. \ref{cmc_curve}, in the three subsets of VehicleID dataset, the CMC curves of DDCL and DDCL (without $\mathcal{L}_P$) are at the top of the CMC curves of $\mathcal{L}_E$, $\mathcal{L}_{SE}$ and $\mathcal{L}_S$, indicating that our proposed DDCL has completely exceeded the $\mathcal{L}_{SE}$ \cite{wen2016discriminative} in the VehicleID dataset. While in the VeRi-776 dataset, although the CMC curve of DDCL is below the CMC curve of $\mathcal{L}_{SE}$ \cite{wen2016discriminative} at the beginning, it become above the CMC curve of $\mathcal{L}_{SE}$ after rank-14. And the CMC curve of $\mathcal{L}_E$ is far below than others, indicating that DDCL makes a great improvement on $\mathcal{L}_E$.

\subsection{Comparison with the state-of-the-art methods}

\begin{table}[h]
	\caption{DDCL compare with the state-of-the-art methods in VeRi-776 dataset}
	\label{tab5}
	\centering
	\setlength{\tabcolsep}{7mm}
	\begin{tabular}{cccc}
		\hline
		Method &  mAP & CMC1 & CMC5 \\
		\hline
		XVGAN  &0.2465&0.6020&0.7703 \\
		FACT + STR &0.2777&0.6144&0.7878	 \\
		OIFE &0.5143&0.6829&0.8971 \\
		S-CNN + P-LSTM  &0.5832&0.8348&0.9004 \\
		RAM  &0.6150&0.8860&0.9400 \\
		VAMI + STR  &0.6132	&0.8592	&0.9184	 \\
		MTCRO  &0.6159&0.8723	&0.9418	 \\
		QD-DLF   &0.6183&0.8850&0.9446 \\
		PAMAL  &0.4506	&0.7205	&0.8886\\
		DAN + AANET(DAVR)  &0.2635&0.6221&0.7366 \\
		
		DDCL $d_\mathcal{E}=600$ (ours) &\textbf{0.6672}&\textbf{0.8940}&\textbf{0.9524} \\		
		
		\hline	
	\end{tabular}
\end{table}

\noindent \textbf{Comparison in VeRi-776 dataset} In Table \ref{tab5}, we list the experimental results of the state-of-the-art methods in the field of vehicle re-identification, include XVGAN \cite{zhou2017cross}, FACT + STR \cite{liu2016adeep}, OIFE \cite{wang2017orientation}, S-CNN + P-LSTM \cite{shen2017learning}, RAM \cite{liu2018ram}, VAMI + STR \cite{zhou2018aware}, MTCRO \cite{xu2018framework}, QD-DLF \cite{zhu2019vehicle}, PAMAL \cite{tumrani2020partial} and DAN + ATTNet (DAVR) \cite{peng2020cross}, and then, compare them with the proposed DDCL. The unsupervised DAN + ATTNet (DAVR) \cite{peng2020cross} and the FACT + STR \cite{liu2016adeep} which combined the traditional features and deep features performed relatively poorly. While OIFE \cite{wang2017orientation}, S-CNN + P-LSTM \cite{shen2017learning}, RAM \cite{liu2018ram}, VAMI + STR \cite{zhou2018aware}, MTCRO \cite{xu2018framework}, QD-DLF \cite{zhu2019vehicle}, PAMAL \cite{tumrani2020partial} and DAN + ATTNet(DAVR) \cite{peng2020cross} use softmax loss as a supervising tools to train their models, after removing softmax loss, the experimental results of DDCL are exceed that of these methods. We only use the original images in this dataset, the experimental results of DDCL exceed that of the methods of XVGAN \cite{zhou2017cross}, VAMI + STR \cite{zhou2018aware}, DAN + ATTNet(DAVR) \cite{peng2020cross}, which use the Generative Adversarial Networks to generate new images to enrich the dataset. DDCL use the single branch, but its experimental results exceed the methods of OIFE \cite{wang2017orientation}, RAM \cite{liu2018ram}, QD-DLF \cite{zhu2019vehicle}, PAMAL \cite{tumrani2020partial}, which use muti branches. And OIFE \cite{wang2017orientation}, RAM \cite{liu2018ram}, PAMAL \cite{tumrani2020partial} extract the local features or attribute features to combine them with the global features, the proposed DDCL only use the global features, but its experimental results exceed that of the above methods.

\begin{table}
	\caption{DDCL compare with the state-of-the-art methods in VeRi-776 dataset.}\label{tab6}
	\setlength{\tabcolsep}{1.8mm}
	\begin{tabular}{ccccccc}
		\hline
		\cline{2-7}
		~ & CMC1 & CMC5 & CMC1 & CMC5 & CMC1 & CMC5 \\
		\hline
		
		GSTE &0.6046	&0.8013	&0.5212	&0.7492	&0.4536	&0.6650	\\
		FDA-Net &0.6403&0.8280&0.5782	&0.7834	&0.4943	&0.7048	\\
		CCL &0.4898&0.7352&0.4280	&0.6685	&0.3822	&0.6167	\\
		XVGAN &0.5287&0.8083&0.4955	&0.7173	&0.4489	&0.6655	\\
		VAMI &0.6312&0.8325	&0.5287	&0.7512	&0.4734	&0.7029	\\
		TAMR &0.6602&0.7971&0.6290	&0.7680	&0.5969	&0.7387	\\
		PAMAL &0.6771&0.8790&0.6150&0.8277	&0.5451	&0.7729	\\
		DAN + ATTNet(DAVR) &0.4948	&0.6866	&0.4518	&0.6399	&0.4071	&0.5902	\\
		DDCL $d_\mathcal{E}=4000$ (ours)&\textbf{0.7320}&\textbf{0.9401}&\textbf{0.6599}&\textbf{0.8951}&\textbf{0.6203}&\textbf{0.8566}	\\

		\hline

	\end{tabular}
\end{table}

\noindent \textbf{Comparison in VehicleID dataset} In this dataset, we compare DDCL with the methods of GSTE \cite{bai2018group}, FDA-Net \cite{lou2019veri}, CCL \cite{liu2016deep}, XVGAN \cite{zhou2017cross}, VAMI \cite{zhou2018aware}, TAMR \cite{guo2019two},  PAMAL \cite{tumrani2020partial} and DAN + ATTNet(DAVR) \cite{peng2020cross}. It is observed from Table \ref{tab6} that the experimental results of DDCL exceed that of the methods of GSTE \cite{bai2018group},  FDA-Net \cite{lou2019veri},  VAMI \cite{zhou2018aware}, TAMR \cite{guo2019two}, PAMAL \cite{tumrani2020partial}, DAN + ATTNet(DAVR) \cite{peng2020cross}, in which the softmax loss are used. The experimental results of DDCL exceed the methods of FDA-Net \cite{lou2019veri}, XVGAN \cite{zhou2017cross}, VAMI \cite{zhou2018aware}, DAN + ATTNet(DAVR) \cite{peng2020cross}, which use the Generative Adversarial Networks to generate new images to enrich the dataset. CCL \cite{liu2016deep}, TAMR \cite{guo2019two}, PAMAL \cite{tumrani2020partial} use multi branch, the experimental results of DDCL exceed that of the methods the above mentioned. And TAMR \cite{guo2019two}, PAMAL \cite{tumrani2020partial} use the local features or attribute features, DDCL only use global features, but the experimental results of DDCL exceed that of the methods above mentioned.

\section{Conclusion}
\label{sec:5}
\ \ \ \ In this paper, we make an insight analysis on center loss, summarize five shortcomings of center loss, and propose a dual distance center loss (DDCL) to improve these shortcomings. On the basis of original Euclidean distance center, we add the Pearson distance center to the same center, which enhance the intra-class compactness, and strengthen the generalization ability of center loss. By proposing the center isolation loss, we improve the later four shortcomings, especially we solve the shortcoming that the center loss must run with the combination of the softmax loss, and combine with the achievement that we verify the softmax loss and the DDCL are inconsistent in the feature space, which makes the DDCL get rid of the restriction of softmax loss, and provides a new perspective to us to examine the center loss. By adding the Pearson center loss, we enhance the intra-class compactness of center loss, by designing the center isolation loss, we strengthen the inter-class separability, by increasing the participation of center parameters and designing a small sample inhibition factor, we change the optimization strategy, and improve the experimental accuracy. All of these, make the proposed DDCL can run without softmax loss and have a high experimental accuracy. In all the four datasets we used in this paper, the performances of DDCL exceed that of softmax loss, and in the two datasets with larger number of training IDs, the performances of DDCL exceed that of softmax loss combine with Euclidean distance center loss, indicating DDCL can perform well in the dataset with large number of training IDs.

\section*{Acknowledgments}
This work is supported by Shenzhen Key Laboratory of Visual Object Detection and Recognition (No. ZDSYS20190902093015527), National Natural Science Foundation of China (No. 61876051) and deep network based high-performance image object detection research (No. JCYJ20180306172101694)

\section*{Appendix: The proof of why with the increase of $\gamma$ in the definition of $\mathcal{L}_{P}$, the average correlation coefficient in the batch will decrease}
\label{appendix}
\ \ \ \ In the formula \ref{LP}, in the process of back propagation, the derivative of $\mathcal{L}_P$ to $x_i$  and $c_j$ are as follows:

\begin{equation}
	\label{f4}
	\frac{\partial \mathcal{L}_P}{\partial x_i}=-\frac{\gamma}{m}\overline{P}^{\gamma-1}\frac{\partial C(x_i,c_{y_i})}{\partial x_i}
\end{equation}

\begin{equation}
	\label{f3}
	\frac{\partial \mathcal{L}_P}{\partial c_j}=-\frac{\gamma}{m}\overline{P}^{\gamma-1}\sum_{i=1}^{m}\delta (y_i==j)\frac{\partial C(x_i,c_j)}{\partial c_j}
\end{equation}
\noindent where

\begin{equation}
	\label{f2}
	\overline{P}=\frac{1}{m}\sum_{i=1}^{m}P(x_i,c_{y_i})=1-\frac{1}{m}\sum_{i=1}^{m}C(x_i,c_{y_i})=1-\overline{C}
\end{equation}

\begin{equation}
	\label{f1}
	\frac{\partial C(x_i,c_{y_i})}{\partial x_i}=\frac{c_{y_i}-\overline{c_{y_i}}}{\Vert x_i-\overline{x_i}\Vert _2\Vert c_{y_i}-\overline{c_{y_i}}\Vert _2}-\frac{(x_i-\overline{x_i})^T(c_{y_i}-\overline{c_{y_i}})}{\Vert x_i-\overline{x_i}\Vert _2^3\Vert c_{y_i}-\overline{c_{y_i}}\Vert _2}(x_i-\overline{x_i})
\end{equation}

\begin{equation}
	\label{f1}
	\frac{\partial C(x_i,c_j)}{\partial c_j}=\frac{x_i-\overline{x_i}}{\Vert x_i-\overline{x_i}\Vert _2\Vert c_j-\overline{c_j}\Vert _2}-\frac{(x_i-\overline{x_i})^T(c_j-\overline{c_j})}{\Vert x_i-\overline{x_i}\Vert _2\Vert c_j-\overline{c_j}\Vert _2^3}(c_j-\overline{c_j})
\end{equation}

\noindent in the formula \ref{f2}, $\overline{C}$ is the average correlation coefficient between features and their centers in a batch. From formula \ref{f4} and formula \ref{f3} we can observe that $\gamma$ must be bigger than 1, otherwise the $\frac{\partial \mathcal{L}_P}{\partial x_i}$ and $\frac{\partial \mathcal{L}_P}{\partial c_j}$ will be very large in the middle and late of the training process, because at this time the $\overline{P}$ is close to zero.
\begin{figure}
	\includegraphics[width=1\textwidth]{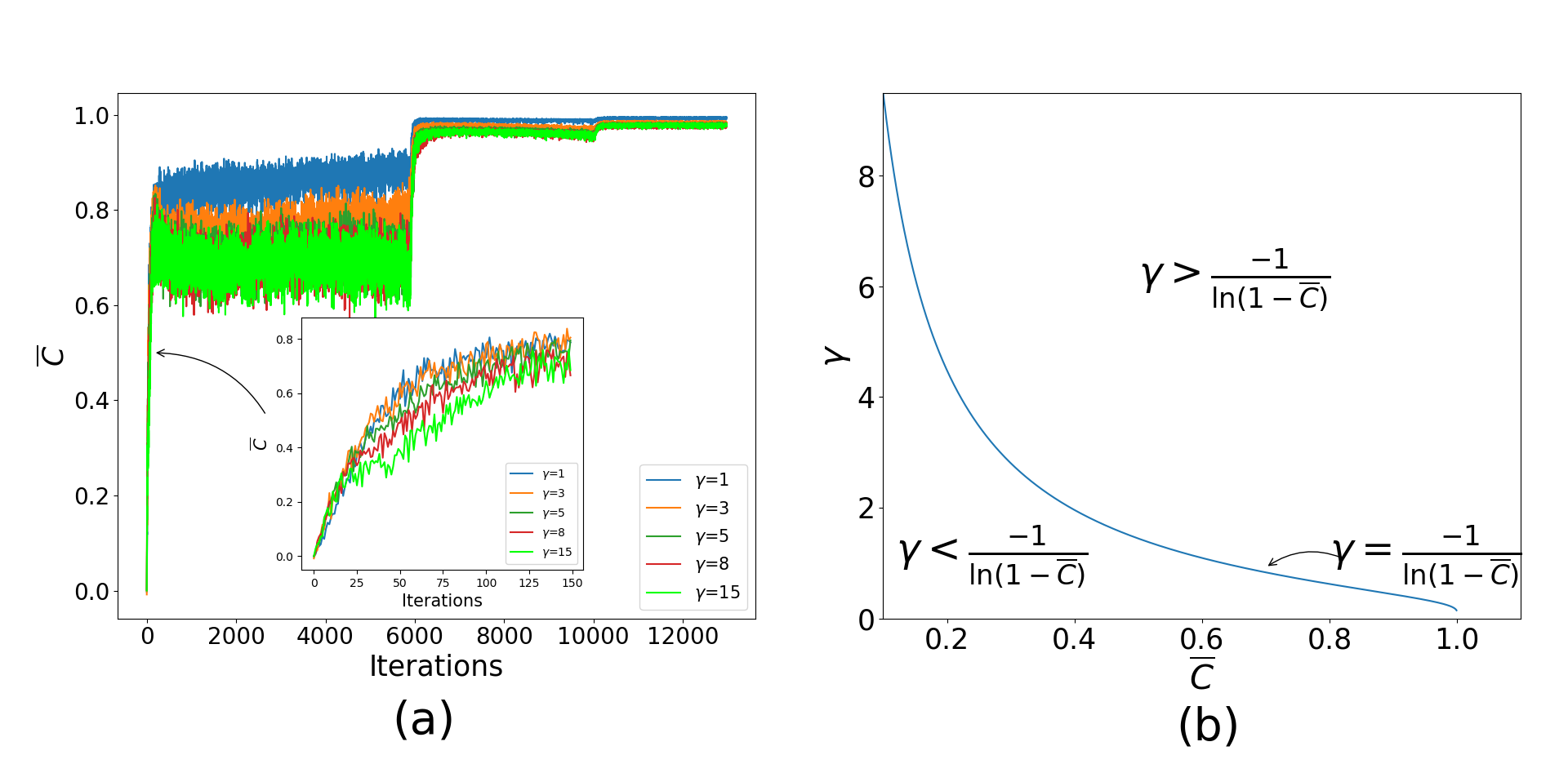}
	\caption{In (a), average correlation coefficient exceeds 0.6 within 150 iterations in VeRi-776 dataset when using $\mathcal{L}_{SEP}$ to supervise training the model. But the total iterations of our training precess in this dataset is 12980. We can consider that the average correlation coefficient is bigger than 0.6 during almost the whole training precess. In (b), when average correlation coefficient is bigger than 0.6, if the $\gamma>1.09$, for each fixed $\overline{C}$, $h(\gamma, \overline{C})$ is an increasing function of  $\gamma$.}
	\label{apendix}       
\end{figure}

During a training epoch, there is only one chance for a feature and its center to move towards each other, but for a center, it has a chance with all the features of the same class to move to each othere, so the position of the center is unknown. Assume the positions of all centers are fixed,  from formula \ref{f4} we know that the value of $\gamma$ effects the value of $\frac{\partial P}{\partial x_i}$ ,in order to analyse this affect, in the formula \ref{f4}, denote
\begin{equation}
	\label{hgc}
	h(\gamma,\overline{C})=\gamma \overline{P}^{\gamma-1}=\gamma (1-\overline{C})^{\gamma-1}
\end{equation}

\noindent when $\overline{C}$ is fixed and $\gamma>1$, we analyse the effect $\gamma$ made to $\frac{\partial L_P}{\partial x_i}$ by discussing the monotonicity of $h(\gamma, \overline{C})$ about $\gamma$ . So

\begin{equation}
	\label{dhgcdg}
	\frac{\partial h(\gamma,\overline{C})}{\partial \gamma}=\left[1+\gamma \ln(1-\overline{C})\right] (1-\overline{C})^{\gamma-1}
\end{equation}

\noindent when $\overline{C}$ is fixed, from formula \ref{dhgcdg} we can obtain that when $\gamma<\frac{-1}{\ln(1-\overline{C})}$, $h(\gamma,\overline{C})$ is a increasing function of $\gamma$ (the region below the curve in Fig. \ref{apendix}), and when $\gamma>\frac{-1}{\ln(1-\overline{C})}$, $h(\gamma,\overline{C})$ is a decreasing function of $\gamma$ (the region above the curve in Fig. \ref{apendix}). Therefor, in the upper region of the curve in the Fig. 1 (b), for every fixed $\overline{C}$, the $h(\gamma,\overline{C})$ is a decreasing function of $\gamma$. From Fig. 1 (b) we know that, when $\overline{C}>0.6$, for every fixed $\overline{C}$, when $\gamma>1.09$,  $h(\gamma,\overline{C})$ is a decreasing function about $\gamma$. For VeRi-776 dataset, we set the batch size to 64, there are 12980 iterations in total, but from Fig. 1 (a) we can observe that, as long as $\gamma>1$, $\overline{C}$ is bigger than 0.6 within 150 iterations during the training process, it makes the  $h(\gamma,\overline{C})$ is a decreasing function of $\gamma$ during almost all the training process. So we get the result that, in the formula \ref{f4}, the larger the $\gamma$, the smaller the absolute value of $\frac{\partial P}{\partial x_i}$, the smaller effect the $\mathcal{L}_P$ makes to the parameters, the bigger the angle between features and their centers, and the smaller the correlation coefficient between features and their centers. The proof is completed.


\bibliographystyle{apalike}
\bibliography{refs} 

\end{document}